\documentclass[11pt]{article}
\usepackage{eamt26}
\usepackage{times}
\usepackage{url}
\usepackage{latexsym}
\usepackage[small,bf]{caption} 
\usepackage{graphicx}
\usepackage[table,xcdraw]{xcolor}
\usepackage{booktabs}
\usepackage{pifont}
\usepackage{fdsymbol}
\usepackage{tabularx}
\usepackage{multirow} 
\usepackage{longtable} 
\usepackage{enumitem}
\usepackage[normalem]{ulem}
\usepackage{accents}
\usepackage{todonotes}
\usepackage{lipsum}
\usepackage{subcaption}
\usepackage{listings} 

\newcommand{\joyemoji}{\raisebox{-0.2em}{\includegraphics[height=1em]{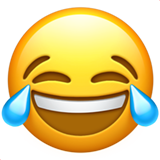}}}

\newcommand{\nllbfull}{\mbox{NLLB-3B}}
\newcommand{\llama}{\mbox{LLaMA}}
\newcommand{\llamafull}{\mbox{LLaMA-3.1-8B}}

\newcommand{\gemmafull}{\mbox{Gemma-2-9B}}
\newcommand{\tower}{\mbox{Tower}}
\newcommand{\towerfull}{\mbox{Tower-0.2-7B}}

\newcommand{\qwenfull}{\mbox{Qwen-2.5-7B}}

\newcommand{\mistralfull}{\mbox{Mistral-0.3-7B}}

\newcommand{\granitefull}{\mbox{Granite-4.1-8B}}
\newcommand{\xtower}{\mbox{xTower}}
\newcommand{\chatgpt}{\mbox{ChatGPT}}
\newcommand{\gpt}{\mbox{GPT}}
\newcommand{\vllm}{vLLM}

\newcommand{\foursquare}{\mbox{Foursquare}}
\newcommand{\mmtc}{\mbox{MMTC}}
\newcommand{\mtnt}{\mbox{MTNT}}

\newcommand{\pfsmb}{\mbox{PFSMB}}
\newcommand{\pmumt}{\mbox{PMUMT}}
\newcommand{\footweets}{\mbox{FooTweets}}
\newcommand{\rocsmt}{\mbox{RoCS-MT}}

\newcommand{\comet}{\mbox{COMET}}
\newcommand{\cometfull}{\mbox{COMET-22}}
\newcommand{\cometkiwi}{\mbox{COMET-Kiwi}}
\newcommand{\sacrecomet}{\mbox{SacreCOMET}}

\newcommand{\bleu}{\mbox{BLEU}}
\newcommand{\sacrebleu}{\mbox{SacreBLEU}}

\newcommand{\ugc}{UGC}
\newcommand{\mt}{MT}

\newcommand{\llm}{LLM}
\newcommand{\llms}{LLMs}

\newcommand{\autoeval}{AE}

\definecolor{normalise}{HTML}{DAF2D0} 
\definecolor{transfer}{HTML}{FBE2D5}
\definecolor{copy}{HTML}{F2CEEF}
\definecolor{omit}{HTML}{FFB3B4}
\definecolor{censor}{HTML}{D0D0D0}

\newcommand{\higherscore}{\colorbox{normalise}{$\uparrow$}}
\newcommand{\similarscore}{\colorbox{censor}{$\approx$}}
\newcommand{\lowerscore}{\colorbox{omit}{$\downarrow$}}

\title{When the Gold Standard Isn't Necessarily Standard:\\ Challenges of Evaluating the Translation of User-Generated Content}

\author{Lydia Nishimwe \quad Beno{\^i}t Sagot \quad Rachel Bawden \\
    Inria \\
    Paris, France \\
  {\tt \{firstname.lastname\}@inria.fr}}

\date{}

\begin{document}
\maketitle
\begin{abstract}
User-generated content (UGC) is characterised by frequent use of non-standard language, from spelling errors to expressive choices such as slang, character repetitions, and emojis. This makes evaluating UGC translation challenging: what counts as a ``good'' translation depends on the desired standardness level of the output. To explore this, we examine the human translation guidelines of four UGC datasets, and derive a taxonomy of twelve non-standard phenomena and five translation actions (\textsc{Normalise}, \textsc{Copy}, \textsc{Transfer}, \textsc{Omit}, \textsc{Censor}). Our analysis reveals notable differences in how UGC is treated, resulting in a spectrum of standardness in reference translations. We show that translation scores of large language models are highly sensitive to prompts with explicit UGC translation instructions, and that they improve when they align with the dataset guidelines. We argue that fair evaluation requires both models and metrics to be aware of translation guidelines. Finally, we call for clear guidelines during dataset creation and for the development of controllable, guideline-aware evaluation frameworks for UGC translation.\footnote{\colorbox{omit}{\textbf{TRIGGER WARNING:}} \ugc\ often contains texts that may be considered explicit, offensive, or vulgar. In this paper, we provide some examples containing profanity. 
We limit ourselves to using explicitly only two words: \textit{f**k} and \textit{sh**}.}
\end{abstract}

\begin{figure}[t]
    \centering
    \includegraphics[width=0.9\linewidth]{figures/ugc-translation-example.png}
    \caption{Example of non-standard phenomena in English--French translation with specific actions. The grammatical error is corrected (\textsc{Normalise}), the irregular capitalisation and word elongation are translated into their French equivalents (\textsc{Transfer}), and the repeated punctuation is copied (\textsc{Copy}).}
    \label{fig:translation-actions}
\end{figure}

\textit{What constitutes a ``good'' translation, and how can it be reliably assessed?} Machine translation (\mt) evaluation seeks to answer these fundamental questions. Human evaluation, performed by linguists, translators, or native speakers, traditionally assessed accuracy and fluency \cite{koehn-monz-2006-manual,callison-burch-etal-2007-meta}, with recent approaches also considering criteria such as style, tone, and terminology \cite{lommel-etal-2014-multidimensional,freitag-etal-2021-experts}. However, human annotations are costly and can be very subjective, motivating the need for automatic evaluation (\autoeval). Reference-based \autoeval, which compares \mt\ outputs to human translations, remains the gold standard \cite{agrawal-etal-2024-automatic-metrics}, though reference translations are costly, often scarce, and can vary in quality \cite{freitag-etal-2021-experts}. On the other hand, reference-less \autoeval\ (or quality estimation) predicts translation quality without references. \autoeval\ metrics have evolved considerably, from string-matching approaches such as \bleu\ \cite{papineni-etal-2002-bleu}, which measure surface-level similarity, to neural models such as \comet\,\cite{rei-etal-2020-comet}, which focus on semantic similarity and are trained to align with human judgments. Neural metrics outperform traditional methods in ranking system outputs \cite{mathur-etal-2020-results,specia-etal-2020-findings-wmt} and, as a more recent phenomenon, large language models (\llms) have also demonstrated strong capabilities in reference-based and reference-less \autoeval\ \cite{freitag-etal-2024-llms,zerva-etal-2024-findings}. Despite these advances, evaluating \mt\ remains a complex and active research area, as evidenced by the long-running, annual WMT Metrics Shared Task \cite{callison-burch-etal-2008-meta,lavie-etal-2025-findings}.

User-generated content (\ugc) adds further complexity, as it is characterised by non-standard phenomena \cite{foster-2010-cba,seddah-etal-2012-french,eisenstein-2013-what,baldwin-li-2015-depth,van-der-goot-etal-2018-taxonomy,sanguinetti-etal-2020-treebanking,bawden-sagot-2023-rocs}, including errors (e.g.,~grammatical, typographical, spelling) and literary devices that convey style, sentiment, and informality (e.g.,~acronyms, slang, phonetisation, code-switching, emojis and other marks of expressiveness).
As a result, translating \ugc\ presents unique challenges beyond those found in traditional text translation, raising questions about how much standardness should be preserved in translation: \textit{Which non-standard phenomena should be corrected and normalised? Which should be maintained, and how?} (see Figure~\ref{fig:translation-actions} for an example). Without explicit guidance, even human ``gold'' translations vary in their treatment of such features, making reference-based \autoeval\ more tricky. 

We address two research questions (\textbf{RQs}): \textbf{(1)~What is a ``good'' translation of \ugc?} and \textbf{(2)~How can \ugc\ translation be evaluated fairly?} First, we analyse four translation datasets of English and French \ugc: \rocsmt\ \cite{bawden-sagot-2023-rocs}, \footweets\ \cite{sluyter-gathje-etal-2018-footweets}, \mmtc\ \cite{mcnamee-duh-2022-multilingual} and \pfsmb\ \cite{rosales-nunez-etal-2019-comparison}. We show that they apply different translation guidelines, a consequence of decisions about standardisation being influenced by the intended use case and context. Second, we evaluate six \llms\ on \ugc\ translation, under varying conditions: zero-shot without guidelines, with dataset-specific guidelines, and cross-application of guidelines between datasets. Through this controlled setup, we show that reference-based \autoeval\ is highly sensitive to the underlying annotation standards, that providing explicit guidelines can significantly shift metric scores, and that fair evaluation of \ugc\ translation requires guideline-aware models and metrics.\footnote{Code and guidelines can be found in \url{https://github.com/lydianish-phd/ugc-mt-eval-challenges}} 

\section{Related Work} 

While most work on \ugc\ translation robustness focuses on handling non-standard phenomena in the source text, Bolding et al.~\shortcite{bolding-etal-2023-ask} is one of the few to address non-standardness in the target side as well. They explore the use of \llms\ to explicitly ``\textit{clean noisy translation data},'' framing robustness primarily as the ability to normalise non-standard language. They use \gpt-4 \cite{openai-2024-gpt4} to remove noise from the target side of the \mtnt\ corpus \cite{michel-neubig-2018-mtnt}, producing a cleaned version intended to better evaluate robustness to non-standard input. Their underlying assumption is that ``\textit{an effective NMT model is capable of translating a noisy source sentence into a clean target sentence}.'' While this perspective aligns with many practical applications, it implicitly adopts a fully normalising view of \ugc\ translation, where non-standard phenomena are treated as noise to be eliminated. In contrast, our work does not assume that the appropriate target of \ugc\ translation is necessarily clean or standardised. We study how different datasets encode distinct translation guidelines, resulting in varying degrees of standardness in the reference translations. Rather than cleaning the data, we treat these guidelines as an explicit modelling and evaluation variable, and investigate how to control \mt\ style by prompting \llms\ to follow them---and how \autoeval\ is affected when such guideline differences are ignored.

Early approaches to controlling \mt\ model translation style (e.g.,~formality, politeness, dialect) included appending style-specific labels/tokens to the input sentences, as well as fine-tuning \cite{sennrich-etal-2016-controlling,niu-carpuat-2020-controlling,rippeth-etal-2022-controlling,riley-etal-2023-frmt}. However, these methods typically required retraining or fine-tuning models on style-specific data, limiting their flexibility. In contrast, the ability of \llms\ to generalise across domains and follow instructions makes them particularly well-suited for controllable \mt. Recent studies have shown that prompting \llms\ with contextual cues can effectively steer the style and register of translations without additional training. Examples of guidelines include indicating specific terminology \cite{moslem-etal-2023-domain,lyu-etal-2024-paradigm}, the intended audience and purpose of the translation \cite{yamada-2023-optimizing}, the domain of the text \cite{gao-etal-2024-how}, the desired tone or style \cite{lyu-etal-2024-paradigm,liu-etal-2024-step}, the language variant \cite{fleisig-etal-2024-linguistic}, and the desired grammatical gender \cite{sanchez-etal-2024-gender}. 

In addition to style transfer, \llms\ have been used to improve the accuracy and fluency of translations. They demonstrate impressive zero-shot robustness to \ugc\ translation, and tend to implicitly normalise and correct non-standard phenomena \cite{bawden-sagot-2023-rocs,peters-martins-2024-did,supryadi-etal-2024-empirical,popovic-etal-2024-effects}. Furthermore, Pan et al.~\shortcite{pan-etal-2024-large} showed that \llms\ could learn \mt\ robustness through in-context examples containing synthetic and natural \ugc. Although \llms\ have been applied to correction tasks such as grammatical error correction \cite{coyne-etal-2023-analyzing,fang-etal-2023-chatgpt,maeng-etal-2023-effectiveness,kwon-etal-2023-chatgpt,penteado-perez-2023-evaluating,katinskaia-yangarber-2024-gpt}, spelling error correction \cite{zhang-etal-2023-does,li-etal-2024-c}, and dialectal normalisation \cite{alam-anastasopoulos-2025-large}, models such as \xtower\ \cite{treviso-etal-2024-xtower} show that \llms\ can also be explicitly prompted to jointly translate and correct errors. 

In our work, we use translation guidelines that relate to the style of the translation (e.g.,~\ugc-specific phenomena and terminology) as well as its fluency (e.g.,~grammar, spelling, punctuation). In some cases, we ask the models to ``transfer'' the non-standardness to its equivalent in the target language, or to only expand abbreviations if the result is not unnatural. This is comparable to the \textit{dynamic equivalence} \cite{nida-1969-theory} prompts of Yamada~\shortcite{yamada-2023-optimizing}, who asked \chatgpt\ to translate a Japanese sentence with cultural references into ``something that would be understood in an English-speaking culture''. 

\begin{table*}[!th]
\centering \small
\setlength\tabcolsep{2pt}
\resizebox{0.8\linewidth}{!}{
\begin{tabular}{r@{}lcccc@{}}
\toprule
  && \multicolumn{1}{c}{En--Fr} & \multicolumn{1}{c}{En--De} & \multicolumn{2}{c}{Fr--En} \\
 &Phenomenon & RoCS-MT & FooTweets & MMTC & PFSMB \\ \midrule
 1. \quad &  Grammar & \cellcolor{normalise}\textsc{Normalise} & \cellcolor{normalise}\textsc{Normalise} & \cellcolor{normalise}\textsc{Normalise} & \cellcolor{normalise}\textsc{Normalise} \\
 2. \quad &  Spelling & \cellcolor{normalise}\textsc{Normalise} & \cellcolor{normalise}\textsc{Normalise} & \cellcolor{normalise}\textsc{Normalise} & \cellcolor{normalise}\textsc{Normalise} \\
 3. \quad &  Word elongation (e.g.,~\texttt{gooooaaaalllll}) & \cellcolor{normalise}\textsc{Normalise} & \cellcolor{transfer}\textsc{Transfer} & \cellcolor{transfer}\textsc{Transfer} & \cellcolor{transfer}\textsc{Transfer} \\
 4. \quad &  Capitalisation (e.g.,~\texttt{NOPE, SoRry}) & \cellcolor{normalise}\textsc{Normalise} & \cellcolor{transfer}\textsc{Transfer} & \cellcolor{transfer}\textsc{Transfer} & \cellcolor{transfer}\textsc{Transfer} \\
 5. \quad &  Informal abbreviations (e.g.,~\texttt{gonna, u}) & \cellcolor{normalise}\textsc{Normalise} & \cellcolor{normalise}\textsc{Normalise} & \cellcolor{normalise}\textsc{Normalise} & \cellcolor{transfer}\textsc{Transfer} \\
 6. \quad &  Informal acronyms (e.g.,~\texttt{LOL, TBH})& \cellcolor{normalise}\textsc{Normalise}\textsuperscript{*} & \cellcolor{normalise}\textsc{Normalise}\textsuperscript{*} & \cellcolor{transfer}\textsc{Transfer} & \cellcolor{transfer}\textsc{Transfer} \\
 7. \quad &  Hashtags and subreddits & \cellcolor{copy}\textsc{Copy} & \cellcolor{copy}\textsc{Copy} & \cellcolor{transfer}\textsc{Transfer} & \cellcolor{transfer}\textsc{Transfer}\textsuperscript{**}\\
 8. \quad &  URLs, user IDs, and retweet marks (\texttt{RT}) & \cellcolor{copy}\textsc{Copy} & \cellcolor{copy}\textsc{Copy} & \cellcolor{copy}\textsc{Copy} & \cellcolor{copy}\textsc{Copy} \\
 9. \quad &  Emoticons and emojis & \cellcolor{copy}\textsc{Copy} & \cellcolor{copy}\textsc{Copy} & \cellcolor{copy}\textsc{Copy} & \cellcolor{copy}\textsc{Copy} \\
 10. \quad &  Atypical punctuation & \cellcolor{normalise}\textsc{Normalise} & \cellcolor{copy}\textsc{Copy} & \cellcolor{copy}\textsc{Copy} & \cellcolor{copy}\textsc{Copy} \\
 11. \quad &  Overt profanity (e.g.,~\texttt{fuck}) & \cellcolor{transfer}\textsc{Transfer} & \cellcolor{transfer}\textsc{Transfer} & \cellcolor{transfer}\textsc{Transfer} & \cellcolor{transfer}\textsc{Transfer} \\
 12. \quad &  Self-censored profanity (e.g.,~\texttt{f*ck}) & \cellcolor{normalise}\textsc{Normalise} & \cellcolor{normalise}\textsc{Normalise} & \cellcolor{normalise}\textsc{Normalise} & \cellcolor{transfer}\textsc{Transfer} \\ \bottomrule
\end{tabular}}
\caption[Summary of guidelines for translating non-standard phenomena as used in the creation of four parallel UGC datasets.]{Summary of guidelines for translating non-standard phenomena as used in the creation of four parallel UGC datasets. \textit{*: Acronyms are expanded (e.g.,~\texttt{TBH $\to$ to be honest}) unless doing so would sound unnatural (e.g.,~\texttt{LOL} is, in practice, more used in its abbreviated form than in its full form \texttt{laughing out loud}). **: Hashtags are translated only if they have a grammatical function in the sentence (e.g. \texttt{\#ItAnnoysMeWhen people don't listen when I'm talking.}).} }
\label{tab:translation-guidelines}
\end{table*}

\section{Methodology}

First (\textbf{RQ1}), we extract and analyse the translation guidelines provided in existing \ugc\ datasets, treating them as a proxy for the human preferences that shape reference translations. Second (\textbf{RQ2}), we conduct a controlled experimental study where \llms\ generate translations under different prompting conditions, allowing us to assess the impact of explicit guidelines on \autoeval\ scores.

\subsection{Translation Guidelines as a Proxy for Human Preferences} \label{sec:translating-ugc}
We define a \textit{translation guideline} as a prescribed action to be applied to a given non-standard linguistic phenomenon in the source text. In what follows, we first describe the main types of non-standard phenomena encountered in \ugc, and then outline the possible actions that may be recommended in guidelines.

\subsubsection{UGC Translation Datasets}

To determine the various decisions that are made when translating non-standard text, we analyse four parallel datasets for the translation of \ugc\ from social media sites and discussion forums: \rocsmt\ \cite{bawden-sagot-2023-rocs} for English--French, \footweets\ \cite{sluyter-gathje-etal-2018-footweets} for English--German, and \mmtc\ \cite{mcnamee-duh-2022-multilingual} and \pfsmb\ \cite{rosales-nunez-etal-2019-comparison} for French--English translation.\footnote{
We left out other well-known datasets such as \mtnt\ \cite{michel-neubig-2018-mtnt} and \foursquare\ \cite{berard-etal-2019-machine} because they provide no details of the guidelines (if any) that were given to the human translators.} 
In particular, the \rocsmt\ dataset creation pipeline includes a lexical normalisation step, and it is the normalised versions that were then translated into the other languages.
This was done to ``optimise the quality of the translation and to reduce the arbitrariness that may be introduced when transferring non-standard variation to the target language''. On the other hand, \footweets\ is a corpus of \textit{Twitter} posts about the 2014 FIFA World Cup created with the downstream task of sentiment analysis in mind: they showed a special focus on preserving the informal nature and the sentiment of the \textit{tweets}.

\mmtc\ and \rocsmt\ have the most detailed lists of guidelines (i.e.,~the most phenomena with specific instructions), followed by \footweets. Conversely, \pfsmb\ has broader instructions that are summarised in two sentences: ``Typographic and grammatical errors were corrected in the gold translations but the language register was kept. For instance, idiomatic expressions were mapped directly to the corresponding (e.g.,~\texttt{mdr} has been translated to \texttt{lol}) and letter repetitions are also kept (e.g.,~\texttt{ouiii} has been translated to \texttt{yesss})''.


\subsubsection{Taxonomy of Non-standard Phenomena}

We curate a list of twelve phenomena from the instructions that were given to human translators in the creation of the datasets. We do that specifically based on the \mmtc\ and \rocsmt\ guidelines (which are the most detailed ones): (1)~Grammar;
(2)~Spelling;
(3)~Word elongation or letter repetitions;
(4)~Capitalisation (e.g.,~missing at the beginning of a sentence or a proper noun, using all caps or case swapping for emphasis); 
(5)~Informal abbreviations;
(6)~Informal acronyms;
(7)~Hashtags and subreddits (e.g.,~\texttt{\#WorldCup, r/Nicegirls}); 
(8)~URLs, @-mentions to user IDs, and retweet marks (RT);
(9)~Emoticons and emojis;
(10)~Atypical punctuation (e.g.,~missing or repeated);
(11)~Overt profanity; and
(12)~Self-censored profanity.

\subsubsection{Taxonomy of Actions}
From the guidelines that were given to human translators in the creation of these datasets, we define three major actions (\textsc{Normalise}, \textsc{Copy} and \textsc{Transfer}) to deal with non-standard phenomena while translating \ugc\ (see Figure~\ref{fig:translation-actions} for an example). We also define two more that do not appear in these guidelines, but can be expected from generated \mt\ outputs: \textsc{Omit} and \textsc{Censor}. The five are described below. 

\begin{enumerate}[leftmargin=*]
    \item \textsc{\textbf{Normalise}}: the non-standard phenomenon is omitted or corrected in the source text, thus producing a standard translation. Examples of this are: correcting spelling and grammatical errors, standardising the use of capital letters and punctuation, expanding abbreviations, removing repeated characters, etc. Note that in the case of self-censored profanity, normalising self-censorship means to render the uncensored version: e.g.,~\texttt{f*ck} $\leftrightarrow$ \texttt{fuck}.
    \item \textsc{\textbf{Copy}}: the non-standard phenomenon is copied as it is in the translation. This is usually applied to special words and characters, such as social media entities, punctuation and emojis.
    \item \textsc{\textbf{Transfer}}: the non-standard phenomenon is mapped to an equivalent non-standard phenomenon in the target language: e.g.,~\texttt{LOL (laughing out oud)} $\rightarrow$ \texttt{MDR (mort de rire)}.  This is not to be confused with \textsc{Copy}, which does not perform any changes. For example, if a hashtag (e.g.,~\texttt{\#WorldCup}) is kept intact in the output, it is \textit{copied}, but if it is translated into a hashtag in the target language (e.g.,~\texttt{\#CoupeDuMonde}), it is \textit{transferred}.
    \item \textsc{\textbf{Omit}}: the non-standard phenomenon is ignored or skipped from the translation.
    This is not to be confused with cases where the omission is a result of standardisation (e.g.,~omitting repeated punctuation marks). Instead, the non-standardness is not dealt with at all (e.g.,~skipping a username mention or URL).
    \item \textsc{\textbf{Censor}}: profanity and offensive language (e.g.,~swear words, insults) are replaced with less offensive terms. E.g.~\texttt{fucked up} $\to$ \texttt{made a mistake}. Other examples include strong or potentially triggering words that are used figuratively:~\texttt{only reason\,I haven't \textit{killed} myself after\,that \textit{boring} game is...}\,$\leftrightarrow$ \texttt{only reason I haven't \textit{harmed} myself after that \textit{uninteresting} game is...}
\end{enumerate}

\subsubsection{Final List of Translation Guidelines} \label{sec:translation-guidelines}

Table~\ref{tab:translation-guidelines} summarises how the 12 non-standard phenomena are translated in the datasets using the taxonomy of actions previously defined.\footnote{For the phenomena without explicit mention in the \footweets\ and \pfsmb\ guidelines, we deduce the instructions by a qualitative analysis of the datasets. For example, we search for a word elongated in the source side (e.g.,~through a search for vowel repetitions) and see how it was translated.} We observe that \rocsmt\ and \pfsmb\ represent two ends of a continuum: \rocsmt\ applies the highest degree of normalisation, while \pfsmb\ preserves the most non-standard phenomena.  \pfsmb\ only has 5 out of 12 guidelines in common with \rocsmt, while \footweets\ and \mmtc\ occupy an intermediate position between these two extremes, respectively with 9 and 7 guidelines in common with \rocsmt. For better readability, we categorise the datasets into two groups corresponding to the level of normalisation of their guidelines and refer to them as: (i)~\rocsmt\ (the ``most standardising''), and (ii)~\footweets, \mmtc, and \pfsmb, (the ``least standardising'').

Note that there are some exceptions provided in the guidelines. For example, \rocsmt\ and \footweets\ translators were suggested to normalise informal acronyms (i.e.,~expand them to their full form), provided that doing so would not sound unnatural in the target language. For instance, \texttt{LOL} is more naturally used in its abbreviated form than in its expanded form \texttt{Laughing Out Loud}. It has even become part of informal vocabulary, producing conjugated forms such as \texttt{I totally LOLed during the movie!}. On the handling of hashtags, we inferred that \pfsmb\ seems to translate them when they serve a grammatical function in the sentence, e.g. \texttt{\#CaMeVénèreQuand on m'écoute pas quand je parle.} $\to$ \texttt{\#ItAnnoysMeWhen people don't listen when I'm talking.}

\subsection{Experimental Study}

We evaluate translation models on the four parallel \ugc\ datasets and the effects of incorporating corpus-specific translation guidelines to control the generation of model outputs.\footnote{See Appendix~\ref{appendix:download-links} for the links to the models and toolkits.}

\paragraph{Translation Models} 
We use the state-of-the-art encoder-decoder model \nllbfull\ \cite{nllbteam-2022-no}  as a baseline for evaluating \mt\ performance. We also evaluate six instruction-tuned, decoder-only \llms: \gemmafull\ \cite{gemmateam-2024-gemma2}, \granitefull\ \cite{ibmresearch-2026-granite4.1}, \llamafull\ \cite{llamateam-2024-llama}, \mistralfull\ \cite{mistralaiteam-2024-mistral7b}, \qwenfull\ \cite{qwenteam-2024-qwen2.5} and \towerfull\ \cite{alves-etal-2024-tower}.
Note that the \tower\ model has been specifically fine-tuned for translation tasks. We generate outputs with a beam search of 5 for \nllbfull\ and we use the \vllm\ toolkit \cite{kwon-2023-efficient} for \llm\ inference, with the following parameters: greedy sampling with BF16 mixed precision \cite{kalamkar-etal-2019-study}, and a maximum model context length of 2,048 tokens. And we prompt the \llms\ to translate one line of text at a time. We also run a post-processing script to extract the translated sentences from verbose outputs and identify refusals to translate \cite{briakou-etal-2024-implications}. The maximum output sequence length is set to 512 tokens for all models. 

\paragraph{Controlled Generation}
In order to control the translation outputs of \llms\ to match the style of a specific corpus's human references, we use a list of 12 translation guidelines derived from Table~\ref{tab:translation-guidelines} as instructions in the \llm\ prompts. Appendix~\ref{appendix:corpus-prompts} details the \llm\ prompt templates and the list of translation guidelines for each corpus.
We define a prompting configuration as a pair constituted of a model and a set of translation guidelines. We evaluate different prompting configurations for each \llm: one without any translation guidelines (the default), and one configuration for each of the specific translation guidelines for each corpus. In particular, we will compare two evaluation scenarios for each \llm\ and dataset:
\begin{enumerate}[leftmargin=*]
    \item \textbf{Matching guidelines}: the guidelines used in the prompts correspond to those originally defined for the dataset, e.g.,~using \rocsmt\ guidelines when translating \rocsmt\ texts;
    \item \textbf{Mismatching guidelines}: the guidelines used in the prompts are taken from a different dataset, e.g.,~using \pfsmb\ guidelines when translating \rocsmt\ texts.
\end{enumerate}

\paragraph{Evaluation Metrics} 
We assess translation quality using neural semantic-based metrics for \autoeval, specifically the reference-based \cometfull\ \cite{rei-etal-2022-comet} and the reference-less \cometkiwi\ \cite{rei-etal-2022-cometkiwi}. 
Following the \sacrecomet\ recommendations \cite{zouhar-etal-2024-pitfalls}, we set sentence-level scores to zero for empty model outputs.
We also use the surface-level metric \bleu\ \cite{papineni-etal-2002-bleu} to complement \cometfull's semantic-based scores.\footnote{We use \sacrebleu\ \cite{post-2018-call} with the signature \texttt{nrefs:1|case:mixed|eff:no|tok:13a| smooth:exp|version:2.6.0} for \bleu\ and \texttt{Python:3.12|PyTorch:2.10.0|version:2.2.7} for the \comet\ models.}
For better readability, we report all scores as percentages, and refer to \cometfull\ simply as \comet.
We evaluate statistical significance using paired bootstrap resampling \cite{koehn-2004-statistical} with 300 samples and a sampling ratio of 0.4. For each metric, we compute the distribution of score differences between each configuration (model + guidelines) and the default no-guideline model. We report the mean difference, 95\% confidence intervals, and $p$-values. A result is considered statistically significant if the 95\% confidence interval of the difference does not include zero.
For the metrics computed at the sentence level (\comet\ and \cometkiwi), bootstrap resampling is performed over sentence-level scores. Corpus-level scores are obtained by averaging the sampled sentence scores.

\paragraph{Human Evaluation}
We conducted a human evaluation on two \ugc\ translation datasets: \rocsmt\ (English--French) and \pfsmb\ (French--English). Specifically, for \pfsmb, we used the \pmumt\ subset \cite{rosales-nunez-etal-2021-understanding}, which was manually normalised and annotated for \ugc\ phenomena. To obtain a unified sampling framework across datasets, we first mapped the original dataset-specific \ugc\ taxonomies to our proposed 12-category taxonomy. We then performed stratified sampling over the mapped categories to ensure coverage of diverse \ugc\ phenomena, resulting in 50 sampled sentence pairs per dataset. We used the outputs of the \gemmafull\ and \llamafull\ models. For each sample, annotators were presented with the source sentence, its normalised version, the reference translation, the applicable guidelines, and two anonymised system outputs corresponding to the default and matching-guideline-based prompting conditions. The system pairs were randomly selected from the outputs of the two models. Annotators were asked to perform pairwise comparisons assessing both overall translation quality and adherence to the \ugc\ guidelines. A total of four annotators, fluent in both English and French, participated in the study. Two annotators annotated both datasets, while two additional annotators annotated only the dataset for which the target language was their native language. Thus, each sample was annotated by three annotators. We measure inter-annotator agreement using Krippendorff's $\alpha$ \cite{krippendorff-1970-estimating} and pairwise Cohen's $\kappa$ \cite{cohen-1960-coefficient}, and obtain final preferences using majority voting over the three annotations. To assess the agreement between human judgments and automatic metrics, we derive pairwise preferences from \comet\ and \cometkiwi\ by comparing the score difference $\Delta = s_{\text{guided}} - s_{\text{default}}$ against a threshold $\epsilon = 0.5$. Differences within $\pm\epsilon$ points are treated as ties.

\begin{table*}[!t]
\centering
\scriptsize
\begin{tabular}{llllll|llll}
\toprule
 &  & \multicolumn{4}{c|}{COMET} & \multicolumn{4}{c}{COMET-Kiwi} \\
Model & Guideline & RoCS-MT & FooTweets & MMTC & PFSMB & RoCS-MT & FooTweets & MMTC & PFSMB \\
\midrule
NLLB-3B & — & 73.21 & 79.90 & 86.54 & 73.17 & 75.52 & 76.51 & 79.52 & 69.83 \\
Gemma-2-9B & None & 76.36 & 83.22 & 77.78 & 78.88 & 77.89 & 79.41 & 77.65 & 75.17 \\
 & +RoCS-MT & \textbf{\underline{76.57}}\higherscore & 84.92\higherscore* & 85.48\higherscore* & 80.29\higherscore* & \underline{77.94}\higherscore & \textbf{\underline{79.97}}\higherscore* & 80.17\higherscore* & \textbf{\underline{75.48}}\higherscore \\
 & +FooTweets & 75.51\lowerscore* & 85.60\higherscore* & 86.44\higherscore* & 80.50\higherscore* & 77.49\lowerscore* & 79.68\higherscore & 80.36\higherscore* & 75.10\lowerscore \\
 & +MMTC & 75.67\lowerscore* & 85.27\higherscore* & \underline{86.59}\higherscore* & \textbf{\underline{80.64}}\higherscore* & 77.55\lowerscore & 79.89\higherscore* & \underline{80.48}\higherscore* & 75.24\higherscore \\
 & +PFSMB & 74.54\lowerscore* & \textbf{\underline{86.20}}\higherscore* & 86.24\higherscore* & 80.51\higherscore* & 77.06\lowerscore* & 78.81\lowerscore* & 80.14\higherscore* & 74.95\lowerscore \\
Granite-4.1-8B & None & 74.27 & 83.66 & 77.56 & 77.95 & 76.84 & \underline{78.65} & 77.66 & 74.11 \\
 & +RoCS-MT & \underline{74.53}\higherscore & 83.71\higherscore & 84.77\higherscore* & 78.18\higherscore & 76.81\lowerscore & 78.46\lowerscore & 79.78\higherscore* & \underline{74.64}\higherscore \\
 & +FooTweets & 74.33\higherscore & \underline{84.12}\higherscore* & 86.44\higherscore* & \underline{78.35}\higherscore & \underline{76.85}\higherscore & 78.41\lowerscore & 80.30\higherscore* & 74.29\higherscore \\
 & +MMTC & 74.16\lowerscore & 83.74\higherscore & 87.77\higherscore* & 78.34\higherscore & 76.75\lowerscore & 78.44\lowerscore & 80.73\higherscore* & 74.28\higherscore \\
 & +PFSMB & 73.75\lowerscore & 83.88\higherscore & \underline{87.99}\higherscore* & 78.32\higherscore & 76.51\lowerscore & 78.19\lowerscore* & \underline{80.76}\higherscore* & 74.04\lowerscore \\
LLaMA-3.1-8B & None & \underline{73.49} & \underline{80.96} & 77.01 & 73.64 & \underline{75.42} & \underline{76.73} & 76.08 & 70.08 \\
 & +RoCS-MT & 71.90\lowerscore* & 75.11\lowerscore* & 81.25\higherscore* & 74.14\higherscore & 73.52\lowerscore* & 70.34\lowerscore* & 76.89\higherscore & \underline{70.32}\higherscore \\
 & +FooTweets & 71.27\lowerscore* & 76.33\lowerscore* & \underline{84.05}\higherscore* & \underline{74.36}\higherscore & 73.09\lowerscore* & 71.29\lowerscore* & \underline{77.76}\higherscore* & 69.99\lowerscore \\
 & +MMTC & 70.94\lowerscore* & 75.19\lowerscore* & 82.72\higherscore* & 71.79\lowerscore & 72.60\lowerscore* & 70.62\lowerscore* & 76.46\higherscore & 67.50\lowerscore* \\
 & +PFSMB & 71.04\lowerscore* & 77.59\lowerscore* & 83.38\higherscore* & 73.07\lowerscore & 73.10\lowerscore* & 72.61\lowerscore* & 77.15\higherscore & 68.45\lowerscore \\
Mistral-0.3-7B & None & 72.33 & 81.18 & 78.46 & 75.73 & 75.00 & 77.02 & 77.17 & 73.22 \\
 & +RoCS-MT & \underline{72.96}\higherscore* & 81.71\higherscore* & 83.43\higherscore* & 76.06\higherscore & \underline{75.28}\higherscore & \underline{77.61}\higherscore* & 78.89\higherscore* & 73.83\higherscore \\
 & +FooTweets & 72.85\higherscore* & 81.99\higherscore* & 84.23\higherscore* & 76.36\higherscore* & \underline{75.28}\higherscore & 77.56\higherscore* & 79.19\higherscore* & \underline{74.04}\higherscore* \\
 & +MMTC & 72.57\higherscore & \underline{82.00}\higherscore* & 84.61\higherscore* & 76.55\higherscore* & 75.08\higherscore & 77.41\higherscore* & 79.24\higherscore* & 73.79\higherscore* \\
 & +PFSMB & 72.47\higherscore & 81.97\higherscore* & \underline{85.31}\higherscore* & \underline{76.64}\higherscore* & 75.14\higherscore & 77.43\higherscore* & \underline{79.49}\higherscore* & 73.86\higherscore* \\
Qwen-2.5-7B & None & \underline{72.19} & 82.70 & 87.10 & 78.67 & \underline{75.25} & \underline{76.83} & \underline{80.39} & \underline{73.75} \\
 & +RoCS-MT & 71.61\lowerscore & \underline{83.55}\higherscore* & 88.39\higherscore* & \underline{78.72}\lowerscore & 74.30\lowerscore* & 75.67\lowerscore* & \underline{80.39} & 73.35\lowerscore \\
 & +FooTweets & 70.48\lowerscore* & 83.46\higherscore* & 88.20\higherscore* & 78.36\lowerscore & 73.43\lowerscore* & 74.91\lowerscore* & 79.92\lowerscore* & 72.53\lowerscore* \\
 & +MMTC & 70.76\lowerscore* & 82.52\lowerscore & \underline{88.42}\higherscore* & 78.63\lowerscore & 73.77\lowerscore* & 75.93\lowerscore* & 80.05\lowerscore & 72.58\lowerscore* \\
 & +PFSMB & 70.60\lowerscore* & 82.81\higherscore & 88.18\higherscore* & 78.33\lowerscore & 73.80\lowerscore* & 75.72\lowerscore* & 80.01\lowerscore & 72.38\lowerscore* \\
Tower-0.2-7B & None & 75.33 & 85.25 & \textbf{\underline{88.94}} & 79.05 & \textbf{\underline{78.16}} & 78.11 & \textbf{\underline{81.04}} & \underline{74.37} \\
 & +RoCS-MT & \underline{75.42}\higherscore & 85.84\higherscore* & 88.87\lowerscore & 79.18\higherscore & 77.92\lowerscore & 78.21\higherscore & 80.93\lowerscore & 74.23\lowerscore \\
 & +FooTweets & 75.35\lowerscore & \underline{85.85}\higherscore* & 88.92\lowerscore & \underline{79.32}\higherscore & 77.88\lowerscore & \underline{78.32}\higherscore & 80.88\lowerscore & 74.30\lowerscore \\
 & +MMTC & 75.31\lowerscore & 85.82\higherscore* & \textbf{\underline{88.94}} & 79.03\lowerscore & 77.90\lowerscore & 78.12\higherscore & 80.94\lowerscore & 74.13\lowerscore \\
 & +PFSMB & 75.26\lowerscore & 85.82\higherscore* & 88.86\lowerscore & 79.04\higherscore & 77.89\lowerscore & 77.97\lowerscore & 80.92\lowerscore & 74.21\lowerscore \\
\bottomrule
\end{tabular}
\caption{COMET and COMET-Kiwi scores for translating \ugc\ with and without corpus-specific guidelines. \textit{Compared to the no-guideline baseline within each model family, score improvements are marked by \higherscore\ and decreases by \lowerscore. Statistical significance with respect to the no-guideline baseline is indicated by * ($95\%$ confidence interval), while the best score within each model family is \underline{underlined}. The best overall score for each dataset is shown in \textbf{bold}.}}
\label{tab:comet-cometkiwi}
\end{table*}

\section{Results and Analysis}

\subsection{Reference-based Quantitative Results}

Table~\ref{tab:comet-cometkiwi} illustrates the \comet\ and \cometkiwi\ scores for evaluating \ugc\ translation across eleven prompting configurations and four datasets. \bleu\ scores are in the appendix Table~\ref{tab:bleu}.

\subsubsection{No-guideline Results} \label{sec:results_sec:no-guidelines}

When prompted without any specific guidelines, \towerfull\ performs the best on all datasets except \rocsmt, where it lags behind \gemmafull. A qualitative analysis of the generated outputs shows that \towerfull's outputs tend to be more formal than \gemmafull's on \rocsmt. Overall, most \llms\ outperform the \nllbfull\ baseline across datasets. The main exception is \mmtc, where \nllbfull\ retains a substantial advantage over most models (up to $\approx10$ \comet\ points in Table~\ref{tab:comet-cometkiwi}). The performance gap on \mmtc\ is likely due to the dataset containing many \textit{Twitter} posts starting with lists of username mentions, which several \llms\ tend to ignore (an instance of the \textsc{Omit} strategy discussed in \S\ref{sec:translation-guidelines}), while \nllbfull, \qwenfull\ and \towerfull\ generally preserve them.\footnote{See the appendix Table~\ref{tab:online-entities} for the number of social media entities in the datasets and default model outputs.} In contrast, \mistralfull\ and \qwenfull\ underperform compared to \nllbfull\ on \rocsmt.

\subsubsection{LLM Instruction Following}

\paragraph{\llama}
Table~\ref{tab:comet-cometkiwi} shows that \llamafull\ results in the lowest \comet\ scores among the evaluated \llms. Furthermore, its performance generally worsens when prompted with translation guidelines (e.g.,~up to a $6$-point drop on \footweets). One reason behind \llamafull's poor performance is that the model refuses to generate translations when it considers the content to be offensive, explicit, hateful or harmful, a behaviour highlighted by \newcite{qian-etal-2024-large-language}. It also fails to produce translations for texts that are too short or seem incomplete (e.g.,~stand-alone hashtags and usernames). The appendix Figure~\ref{fig:llama-censorship} illustrates the percentage of  translation requests refused by \llamafull. We observe the highest percentage for the no-guideline scenario ($3\%$)  on \pfsmb. Moreover, adding corpus-specific guidelines significantly increases the ratio of refused requests (up to $8\%$ on \pfsmb\ and \mmtc), likely because all guidelines include explicit instructions to preserve profanity in the translation. In contrast, we notice lower rates of refusal on \footweets\ and \rocsmt, even with the guidelines ($\le 4\%$), possibly because \footweets\ (based on football reactions) is more likely to have ``safer'' content. Meanwhile, explicit or triggering content are specifically filtered out in the \rocsmt\ dataset creation pipeline. 

\paragraph{\tower}
\towerfull, which was built on top of a \llama-2 model \cite{touvron-etal-2023-llama2}, does not display the same censored behaviour as \llamafull. However, Table~\ref{tab:comet-cometkiwi} shows that prompting \towerfull\ with translation guidelines results in minimal gains compared to the no-guideline scenario. These results suggest that \towerfull\ tends to stick to its own translation style, ignoring the instructions. To further confirm this, we compute \bleu\ scores to measure the lexical overlap between the model outputs of all configurations (with and without guidelines) and report the scores for \gemmafull\ and \towerfull\ in the appendix Figure~\ref{fig:gemma-tower}. We see little lexical variation between \towerfull's outputs, regardless of translation guidelines.

\paragraph{Other models}
In contrast, the other evaluated \llms\ (\gemmafull, \granitefull, \mistralfull\ and \qwenfull) produce more noticeable variations across guideline configurations and, unlike \llamafull, they do not exhibit systematic refusal behaviour. This suggests that these models are more willing to adapt their translation strategies to the requested \ugc\ handling instructions, an effect that we analyse further in the following sections.

\subsubsection{Effects of Corpus-specific Guidelines} \label{sec:results_sec:with-guidelines}

The following observations focus on \gemmafull, \mistralfull, \granitefull\ and \qwenfull. We note that when restricting \llamafull\ to the subset of samples for which the model does not refuse to translate, we observe the same overall trends.

\paragraph{Matching guidelines on all datasets}
Across most models and datasets, adding corpus-specific translation guidelines improves \comet\ scores compared to the no-guideline setting, often with statistically significant gains (Table~\ref{tab:comet-cometkiwi}). In many cases, the best results are obtained with the matching dataset guidelines, particularly for \gemmafull, \granitefull\ and \mistralfull. Overall, \gemmafull\ with guidelines achieves the best results across datasets, except on \mmtc, where the default \towerfull\ model remains the strongest system.

\paragraph{Mismatching guidelines between the least standardising datasets}
Applying guidelines from the least standardising datasets (\footweets, \mmtc\ and \pfsmb) generally improves performance across these datasets, often with statistically significant gains. In several cases, preservation-oriented guidelines from one dataset transfer effectively to another preservation-oriented dataset, and the score variations between these guideline configurations remain very small. In particular, guideline prompting substantially improves the handling of social media entities on \mmtc, where several models tend to omit usernames and online entities by default. However, the least standardising guidelines, especially those of \pfsmb, can also degrade translation quality by encouraging models to over-transfer non-standardness. This behaviour is particularly noticeable for \qwenfull, which appears less robust to non-standard source inputs and more prone to generating hallucinated forms, ungrammatical constructions or noisy lexical variations when attempting to follow preservation-oriented instructions. On \pfsmb, guideline prompting leads to statistically significant improvements for \gemmafull\ and \mistralfull, while the variations observed for the other models remain non-significant.

\paragraph{Mismatching guidelines between \rocsmt\ and the least standardising datasets}
In contrast, applying the less standardising guidelines to \rocsmt\ generally results in negligible variations or decreases in performance. Since \rocsmt\ follows the most standardising translation strategy, these results suggest that several models already tend to normalise \ugc\ phenomena by default. Consequently, introducing preservation-oriented instructions often moves the outputs away from the reference style. We observe statistically significant decreases for \gemmafull\ and \qwenfull, while the variations for \granitefull\ and \mistralfull\ remain non-significant. Conversely, applying \rocsmt\ guidelines to the other datasets generally improves performance compared to the default setting, although these gains are usually smaller than those obtained with the less standardising guidelines. Overall, these findings indicate that guideline prompting is most effective when the prompting strategy matches the intended level of normalisation or preservation in the target dataset.

\subsection{Reference-less Quantitative Results}

The \cometkiwi\ results in Table~\ref{tab:comet-cometkiwi} differ from the reference-based \comet\ results in that the best scores are generally obtained with either no guidelines (default \towerfull) or with more standardising guideline configurations (\rocsmt-guided \gemmafull), rather than necessarily with the guidelines matching the dataset. 
However, similarly to \comet, the score variations across guideline configurations remain small and non-significant. Furthermore, when the guidelines lead to less standard outputs, the metric tends to assign lower scores. In particular, the least standardising guidelines, especially those of \pfsmb, frequently result in the lowest \cometkiwi\ scores. This behaviour is particularly noticeable for \qwenfull, whose outputs become substantially noisier under preservation-oriented prompting. 
These results suggest that the metric remains biased towards standardised outputs and is less robust to higher levels of non-standardness. This is consistent with Aepli et al.~\shortcite{aepli-etal-2023-benchmark}'s conclusion that \cometkiwi\ is not robust to non-standard orthographic variation in dialects and is biased towards standardised outputs.

\subsection{Qualitative Analysis}


The appendix Table~\ref{tab:outputs} illustrates a few example sentences from the datasets and their output translations by \gemmafull, with and without matching guidelines, as well as their sentence-level \comet\ scores.
We observe that including specific translation guidelines in the prompt helps \gemmafull\ to better deal with certain \ugc-specific phenomena in a way that differs from its default behaviour, thus increasing the sentence-level scores overall (see examples~1,~2 and~3).

Note, however, that the model's behaviour may be inconsistent:
\gemmafull\ is not always able to apply the guidelines correctly. A prominent example of this is its tendency to translate (\textsc{Transfer}) hashtags on \footweets, despite clear instructions  to \textsc{Copy}, as seen in examples~4 and~5. Other instances include the failure to \textsc{Transfer} character repetitions (e.g.,~\texttt{niceeee} $\to$ \texttt{schööön} in example~4), and some attempts to do so even lead the model to repeat characters indefinitely. One possible reason for this is that this guideline contradicts the instruction to \textsc{Normalise} spelling, raising questions about the compositionality of the twelve defined guidelines and the overall feasibility of applying them consistently. Likewise, instructions to \textsc{copy} irregular punctuation and capitalisation can contradict the guideline to \textsc{Normalise} grammar. 

In addition, \textsc{Transferring} non-standardness can degrade translation quality, as evidenced by the ungrammatical formulation `\texttt{no even need}' in example~6 and the implicit repetition `\texttt{rn now}' in example~7 (which significantly lowered the \comet\ score).\footnote{`\texttt{rn}' stands for `\texttt{right now}'.}
We also observed that the profanity guidelines can cause \gemmafull\ to hallucinate swear words or explicit terms. 

Finally, example~8 shows that, while prompting \gemmafull\ with guidelines helps control the translation output style, it does not increase the model's robustness to the non-standard phenomena that it inherently fails to translate.

\subsection{Human Evaluation Results}

The human evaluation results broadly confirm the trends observed with the automatic metrics (see the appendix Table~\ref{tab:human-eval-full}). On both datasets, the guided outputs are preferred more often than the default outputs for both overall quality and guideline adherence. On \pfsmb, the guided outputs are preferred in $38\%$ of all samples for overall quality, compared to only $12\%$ for the default outputs, while half of the samples are judged as ties. The effect is even stronger for guideline adherence, where the guided outputs are preferred in $36\%$ of the samples compared to only $4\%$ for the default outputs. When ties are ignored, these proportions rise to $76\%$ and $90\%$, respectively. In contrast, the gains on \rocsmt\ are smaller, with guided outputs preferred in $30\%$ of the samples for quality and $12\%$ for guideline adherence, while ties account for $54\%$ and $80\%$ of the judgments. This is consistent with the quantitative results; the models already tend to produce relatively standardised outputs by default, making the impact of additional prompting less noticeable on a dataset whose references are highly standardising.

The joint evaluation outcomes further highlight the differences between the two datasets. On \pfsmb, the guided outputs are preferred on both dimensions in $24\%$ of the evaluated samples, compared to only $10\%$ on \rocsmt. Furthermore, we observe virtually no cases where the guided outputs improve adherence at the expense of translation quality ($0\%$ on \pfsmb\ and $2\%$ on \rocsmt). This indicates that preservation-oriented prompting can improve the handling of \ugc\ phenomena without substantially degrading translation quality.

Inter-annotator agreement is moderate overall (Krippendorff's $\alpha=0.25$--$0.45$). We also observe moderate to substantial agreement between human preferences and automatic metrics. Overall quality judgments show the strongest agreement with \cometkiwi, reaching Cohen's $\kappa$ values of $0.62$ on \rocsmt\ and $0.54$ on \pfsmb. In contrast, guideline adherence judgments are better captured by \comet, particularly on \pfsmb\ ($\kappa = 0.56$ compared to $0.46$ for \cometkiwi). This is consistent with the fact that \comet\ has access to the references and can therefore better capture adherence to the translation style they embody.

\section{Discussion}

\paragraph{Defining the ``gold standard'' for \ugc\ translation (RQ1)}
By analysing the human references of different corpora and how they treat non-standard phenomena, we can infer style preferences for \ugc\ translation. Our study of four datasets shows that what counts as a ``good'' translation varies with the reference style: some datasets are highly standardising (\rocsmt), and others minimally standardising (\pfsmb), with \footweets\ and \mmtc\ in between. Using our 12-phenomena taxonomy and five actions (\textsc{Normalise}, \textsc{Copy}, \textsc{Transfer}, \textsc{Omit}, \textsc{Censor}), we observe systematic differences in how acronyms, hashtags, letter repetitions, and social media entities are handled. Both the automatic and human evaluation results further confirm that these stylistic differences materially affect model performance and human preferences. This demonstrates that translation of \ugc\ is inherently guideline-dependent and context-sensitive.  

\paragraph{Guideline awareness in models and metrics for fair evaluation (RQ2)}
Fair evaluation requires putting models in comparable situations, taking into account the fact that \ugc\ translation is guideline-dependent (see RQ1). Ideally, models should be guided by the same translation principles used to generate references. \nllbfull, an encoder-decoder model that is not style-controlled, cannot be prompted with guidelines and thus produces a stable but standard style. On the other hand, instruction-tuned \llms\ such as \gemmafull, \granitefull, \mistralfull\ and \qwenfull\ adapt their outputs to guideline-based prompting, while \towerfull\ shows minimal sensitivity to the prompts, suggesting limited stylistic controllability. In contrast, \llamafull\ frequently refuses to translate ``unsafe'' inputs, lowering corpus-level scores and complicating comparisons. 
We also observe important differences between evaluation metrics. Reference-based metrics (\bleu, \comet) are implicitly style-aware, rewarding outputs aligned with reference guidelines, but penalising mismatches, as with the least-standardising prompts applied to \rocsmt. In contrast, the reference-less \cometkiwi\ metric tends to favour more standardised outputs, assigning lower scores to highly non-standard translations even when they better match the intended preservation-oriented guidelines, as seen with \pfsmb.
This highlights the challenge of reliably scoring extreme variation without guideline alignment.

\paragraph{Recommendations}
We make two practical recommendations for ensuring a fairer evaluation of \ugc\ translation:
(1) when style is not a priority and all linguistic variations should be treated equally, use either a reference-less metric or a reference-based metric with multiple versions of the references spanning different levels of standardness;
(2) when control over a specific style is required, follow the approach proposed in this work by prompting an \llm\ with explicit guidelines and evaluating outputs with reference-based methods, ideally an \llm-as-a-judge \cite{zheng-etal-2023-judging}  configured with the same guidelines to provide a controllable, style-sensitive evaluation.
More generally, our results suggest that future work on \ugc\ evaluation would benefit from more fine-grained human evaluation protocols explicitly targeting individual \ugc\ phenomena and translation actions. Rather than relying solely on overall quality judgments, future evaluation frameworks could include phenomenon-specific rubrics.

\section{Conclusion}

In this work, we investigated the role of translation guidelines in the evaluation of \ugc\ machine translation. Through the analysis of four datasets, we showed that \ugc\ translation is inherently style-dependent, with datasets reflecting different translation philosophies ranging from highly standardising to preservation-oriented approaches. Using a unified taxonomy of \ugc\ phenomena and translation actions, we highlighted systematic differences in how non-standard language is treated across corpora.

Our experiments further demonstrated that modern \llms\ differ substantially in their ability to follow stylistic translation instructions. While some models successfully adapt to corpus-specific prompting strategies, others either ignore the instructions or exhibit unstable behaviour when handling highly non-standard or explicit content. Both the automatic and human evaluation results show that guideline prompting is most effective when the prompting strategy matches the stylistic assumptions underlying the reference data.

Finally, our findings highlight important limitations of current evaluation practices for \ugc\ translation. Reference-based metrics are strongly influenced by the stylistic properties of the references, while reference-less metrics tend to favour more standardised outputs and remain less robust to highly non-standard language. Overall, our results emphasise the need for more controllable, guideline-aware evaluation frameworks capable of accounting for multiple valid translation styles and varying levels of non-standardness in \ugc.

\section*{Limitations}

\paragraph{Implicit vs.~Explicit Translation Guidelines} We aimed to use explicit annotation guidelines as a proxy for human preferences. Inferring missing explicit guidelines from the implicit annotator choices can be seen as introducing circularity. However, we argue that in such cases, the guideline creators delegate the decision to the annotators, thus letting the latter’s preferences take precedence over their own. In other words: explicit guidelines reflect the preferences of the creators, while implicit ones reflect those of the annotators.

\paragraph{Guideline Descriptions} Some guidelines were too vague, contradictory, or inconsistently defined to be applied reliably. Our taxonomy and instruction set may also omit certain phenomena or lack sufficient granularity. A more exhaustive and clearly structured set of guidelines, potentially with example-based definitions per phenomenon, could improve both annotation consistency and model interpretability. 

\paragraph{Prompt Engineering} We do not explore a wide range of prompting strategies or formulations of the translation guidelines. In particular, our use of zero-shot prompts without few-shot demonstrations may have limited the models’ ability to fully adhere to the guidelines  \cite{hendy-etal-2023-good,garcia-etal-2023-unreasonable,coyne-etal-2023-analyzing,gao-etal-2024-how,sclar-etal-2024-quantifying,ceballos-arroyo-etal-2024-open,chatterjee-etal-2024-posix,pan-etal-2024-large,zebaze-etal-2025-compositional,zebaze-etal-2025-context}. Future work could investigate how different prompt structures or the inclusion of targeted few-shot examples affect model behaviour in handling specific non-standard phenomena. In particular, it might prove beneficial to use chain-of-thought prompting to instruct the model to handle different subsets of the guidelines sequentially, e.g. first correcting grammar and spelling, and then re-injecting non-standardness in the corrected version.

\paragraph{Metric Coverage} We only evaluated using \comet, \cometkiwi, and \bleu. Including additional metrics, particularly other neural-based and \llm-based evaluation methods, would allow for a more comprehensive comparison and help generalise the conclusions regarding the sensitivity of metrics to \ugc\ style and guideline adherence.

\section*{Acknowledgements}
We thank the reviewers for their constructive feedback. We also thank Marine Carpuat and Armel Zebaze for helpful discussions on experiment design, Jakob Maier for qualitative analysis of German translations, Rasul Dent for French--English human evaluation, and Alfred Buregeya for proofreading an earlier draft. This work was granted access to the HPC resources of IDRIS under the allocations 2023AD011013674R2 made by GENCI. It was  partly funded by Rachel Bawden and Benoît Sagot’s chairs in the PRAIRIE institute, funded by the French national agency ANR, as part of the “Investissements d’avenir” programme under the reference ANR-19-P3IA-0001 and by Benoît Sagot's chair in its follow-up, PRAIRIE-PSAI, also funded by the ANR as part of the “France 2030” strategy under the reference ANR23-IACL-0008. It also received further funding from the ANR under the project TraLaLaM (``ANR-23-IAS1-0006'').

\bibliography{biblio}

@inproceedings{koehn-monz-2006-manual,
    title = "{Manual and Automatic Evaluation of Machine Translation between European Languages}",
    author = "Koehn, Philipp  and
      Monz, Christof",
    xeditor = "Koehn, Philipp  and
      Monz, Christof",
    booktitle = "Proceedings on the Workshop on Statistical Machine Translation",
    xmonth = jun,
    year = "2006",
    address = "New York City",
    publisher = "Association for Computational Linguistics",
    url = "https://aclanthology.org/W06-3114/",
    pages = "102--121"
}

@inproceedings{callison-burch-etal-2007-meta,
    title = "{(Meta-) Evaluation of Machine Translation}",
    author = "Callison-Burch, Chris  and
      Fordyce, Cameron  and
      Koehn, Philipp  and
      Monz, Christof  and
      Schroeder, Josh",
    xeditor = "Callison-Burch, Chris  and
      Koehn, Philipp  and
      Fordyce, Cameron Shaw  and
      Monz, Christof",
    booktitle = "Proceedings of the Second Workshop on Statistical Machine Translation",
    xmonth = jun,
    year = "2007",
    address = "Prague, Czech Republic",
    publisher = "Association for Computational Linguistics",
    url = "https://aclanthology.org/W07-0718/",
    pages = "136--158"
}

@article{lommel-etal-2014-multidimensional,
      author = {Arle Lommel and Hans Uszkoreit
               and Aljoscha Burchardt},
       title = {{Multidimensional Quality Metrics (MQM): a Framework for
               Declaring and Describing Translation Quality Metrics}},
     journal = {Tradumàtica},
        year = {2014},
      volume = {12},
       pages = {455--463},
         doi = {10.5565/rev/tradumatica.77},
         url = {https://ddd.uab.cat/record/130144},
}

@article{freitag-etal-2021-experts,
    title = "{Experts, Errors, and Context: A Large-Scale Study of Human Evaluation for Machine Translation}",
    author = "Freitag, Markus  and
      Foster, George  and
      Grangier, David  and
      Ratnakar, Viresh  and
      Tan, Qijun  and
      Macherey, Wolfgang",
    xeditor = "Roark, Brian  and
      Nenkova, Ani",
    journal = "Transactions of the Association for Computational Linguistics",
    volume = "9",
    year = "2021",
    address = "Cambridge, MA",
    publisher = "MIT Press",
    url = "https://aclanthology.org/2021.tacl-1.87/",
    doi = "10.1162/tacl_a_00437",
    pages = "1460--1474"
}

@inproceedings{agrawal-etal-2024-automatic-metrics,
    title = "{Can Automatic Metrics Assess High-Quality Translations?}",
    author = "Agrawal, Sweta  and
      Farinhas, Ant{\'o}nio  and
      Rei, Ricardo  and
      Martins, Andre",
    xeditor = "Al-Onaizan, Yaser  and
      Bansal, Mohit  and
      Chen, Yun-Nung",
    booktitle = "Proceedings of the 2024 Conference on Empirical Methods in Natural Language Processing",
    xmonth = nov,
    year = "2024",
    address = "Miami, Florida, USA",
    publisher = "Association for Computational Linguistics",
    url = "https://aclanthology.org/2024.emnlp-main.802/",
    doi = "10.18653/v1/2024.emnlp-main.802",
    pages = "14491--14502"
}

@inproceedings{papineni-etal-2002-bleu,
	address = {Philadelphia, Pennsylvania, USA},
	title = {{BLEU}: a {Method} for {Automatic} {Evaluation} of {Machine} {Translation}},
	shorttitle = {Bleu},
    url = {https://aclanthology.org/P02-1040},
	doi = {10.3115/1073083.1073135},
    urldate = {2021-11-03},
	booktitle = {Proceedings of the 40th {Annual} {Meeting} of the {Association} for {Computational} {Linguistics}},
	publisher = {Association for Computational Linguistics},
	author = {Papineni, Kishore and Roukos, Salim and Ward, Todd and Zhu, Wei-Jing},
	xmonth = jul,
	year = {2002},
	pages = {311--318},
}

@inproceedings{rei-etal-2020-comet,
	address = {Online},
	title = {{COMET}: {A} {Neural} {Framework} for {MT} {Evaluation}},
	shorttitle = {{COMET}},
    url = {https://aclanthology.org/2020.emnlp-main.213},
	doi = {10.18653/v1/2020.emnlp-main.213},
    urldate = {2023-03-01},
	booktitle = {Proceedings of the 2020 {Conference} on {Empirical} {Methods} in {Natural} {Language} {Processing} ({EMNLP})},
	publisher = {Association for Computational Linguistics},
	author = {Rei, Ricardo and Stewart, Craig and Farinha, Ana C and Lavie, Alon},
	xmonth = nov,
	year = {2020},
	pages = {2685--2702},
}

@inproceedings{mathur-etal-2020-results,
    title = "{Results of the {WMT}20 Metrics Shared Task}",
    author = "Mathur, Nitika  and
      Wei, Johnny  and
      Freitag, Markus  and
      Ma, Qingsong  and
      Bojar, Ond{\v{r}}ej",
    xeditor = {Barrault, Lo{\"i}c  and
      Bojar, Ond{\v{r}}ej  and
      Bougares, Fethi  and
      Chatterjee, Rajen  and
      Costa-juss{\`a}, Marta R.  and
      Federmann, Christian  and
      Fishel, Mark  and
      Fraser, Alexander  and
      Graham, Yvette  and
      Guzman, Paco  and
      Haddow, Barry  and
      Huck, Matthias  and
      Yepes, Antonio Jimeno  and
      Koehn, Philipp  and
      Martins, Andr{\'e}  and
      Morishita, Makoto  and
      Monz, Christof  and
      Nagata, Masaaki  and
      Nakazawa, Toshiaki  and
      Negri, Matteo},
    booktitle = "Proceedings of the Fifth Conference on Machine Translation",
    xmonth = nov,
    year = "2020",
    address = "Online",
    publisher = "Association for Computational Linguistics",
    url = "https://aclanthology.org/2020.wmt-1.77/",
    pages = "688--725"
}

@inproceedings{specia-etal-2020-findings-wmt,
    title = "{Findings of the {WMT} 2020 Shared Task on Quality Estimation}",
    author = "Specia, Lucia  and
      Blain, Fr{\'e}d{\'e}ric  and
      Fomicheva, Marina  and
      Fonseca, Erick  and
      Chaudhary, Vishrav  and
      Guzm{\'a}n, Francisco  and
      Martins, Andr{\'e} F. T.",
    xeditor = {Barrault, Lo{\"i}c  and
      Bojar, Ond{\v{r}}ej  and
      Bougares, Fethi  and
      Chatterjee, Rajen  and
      Costa-juss{\`a}, Marta R.  and
      Federmann, Christian  and
      Fishel, Mark  and
      Fraser, Alexander  and
      Graham, Yvette  and
      Guzman, Paco  and
      Haddow, Barry  and
      Huck, Matthias  and
      Yepes, Antonio Jimeno  and
      Koehn, Philipp  and
      Martins, Andr{\'e}  and
      Morishita, Makoto  and
      Monz, Christof  and
      Nagata, Masaaki  and
      Nakazawa, Toshiaki  and
      Negri, Matteo},
    booktitle = "Proceedings of the Fifth Conference on Machine Translation",
    xmonth = nov,
    year = "2020",
    address = "Online",
    publisher = "Association for Computational Linguistics",
    url = "https://aclanthology.org/2020.wmt-1.79/",
    pages = "743--764"
}

@inproceedings{freitag-etal-2024-llms,
    title = "{Are {LLM}s Breaking {MT} Metrics? Results of the {WMT}24 Metrics Shared Task}",
    author = "Freitag, Markus  and
      Mathur, Nitika  and
      Deutsch, Daniel  and
      Lo, Chi-Kiu  and
      Avramidis, Eleftherios  and
      Rei, Ricardo  and
      Thompson, Brian  and
      Blain, Frederic  and
      Kocmi, Tom  and
      Wang, Jiayi  and
      Adelani, David Ifeoluwa  and
      Buchicchio, Marianna  and
      Zerva, Chrysoula  and
      Lavie, Alon",
    xeditor = "Haddow, Barry  and
      Kocmi, Tom  and
      Koehn, Philipp  and
      Monz, Christof",
    booktitle = "Proceedings of the Ninth Conference on Machine Translation",
    xmonth = nov,
    year = "2024",
    address = "Miami, Florida, USA",
    publisher = "Association for Computational Linguistics",
    url = "https://aclanthology.org/2024.wmt-1.2/",
    doi = "10.18653/v1/2024.wmt-1.2",
    pages = "47--81"
}

@inproceedings{zerva-etal-2024-findings,
    title = "{Findings of the Quality Estimation Shared Task at {WMT} 2024: Are {LLM}s Closing the Gap in {QE}?}",
    author = "Zerva, Chrysoula  and
      Blain, Frederic  and
      C. De Souza, Jos{\'e} G.  and
      Kanojia, Diptesh  and
      Deoghare, Sourabh  and
      Guerreiro, Nuno M.  and
      Attanasio, Giuseppe  and
      Rei, Ricardo  and
      Orasan, Constantin  and
      Negri, Matteo  and
      Turchi, Marco  and
      Chatterjee, Rajen  and
      Bhattacharyya, Pushpak  and
      Freitag, Markus  and
      Martins, Andr{\'e}",
    xeditor = "Haddow, Barry  and
      Kocmi, Tom  and
      Koehn, Philipp  and
      Monz, Christof",
    booktitle = "Proceedings of the Ninth Conference on Machine Translation",
    xmonth = nov,
    year = "2024",
    address = "Miami, Florida, USA",
    publisher = "Association for Computational Linguistics",
    url = "https://aclanthology.org/2024.wmt-1.3/",
    doi = "10.18653/v1/2024.wmt-1.3",
    pages = "82--109"
}

@inproceedings{callison-burch-etal-2008-meta,
    title = "{Further Meta-Evaluation of Machine Translation}",
    author = "Callison-Burch, Chris  and
      Fordyce, Cameron  and
      Koehn, Philipp  and
      Monz, Christof  and
      Schroeder, Josh",
    xeditor = "Callison-Burch, Chris  and
      Koehn, Philipp  and
      Monz, Christof  and
      Schroeder, Josh  and
      Fordyce, Cameron Shaw",
    booktitle = "Proceedings of the Third Workshop on Statistical Machine Translation",
    xmonth = jun,
    year = "2008",
    address = "Columbus, Ohio",
    publisher = "Association for Computational Linguistics",
    url = "https://aclanthology.org/W08-0309/",
    pages = "70--106"
}

@inproceedings{lavie-etal-2025-findings,
    title = "{Findings of the {WMT}25 Shared Task on Automated Translation Evaluation Systems: Linguistic Diversity is Challenging and References Still Help}",
    author = "Lavie, Alon  and
      Hanneman, Greg  and
      Agrawal, Sweta  and
      Kanojia, Diptesh  and
      Lo, Chi-Kiu  and
      Zouhar, Vil{\'e}m  and
      Blain, Frederic  and
      Zerva, Chrysoula  and
      Avramidis, Eleftherios  and
      Deoghare, Sourabh  and
      Sindhujan, Archchana  and
      Wang, Jiayi  and
      Adelani, David Ifeoluwa  and
      Thompson, Brian  and
      Kocmi, Tom  and
      Freitag, Markus  and
      Deutsch, Daniel",
    xeditor = "Haddow, Barry  and
      Kocmi, Tom  and
      Koehn, Philipp  and
      Monz, Christof",
    booktitle = "Proceedings of the Tenth Conference on Machine Translation",
    xmonth = nov,
    year = "2025",
    address = "Suzhou, China",
    publisher = "Association for Computational Linguistics",
    url = "https://aclanthology.org/2025.wmt-1.24/",
    doi = "10.18653/v1/2025.wmt-1.24",
    pages = "436--483",
    ISBN = "979-8-89176-341-8"
}

@inproceedings{foster-2010-cba,
	address = {Los Angeles, California},
	title = {“cba to check the spelling”: {Investigating} {Parser} {Performance} on {Discussion} {Forum} {Posts}},
	shorttitle = {“cba to check the spelling”},
    url = {https://aclanthology.org/N10-1060},
    urldate = {2021-10-20},
	booktitle = {Human {Language} {Technologies}: {The} 2010 {Annual} {Conference} of the {North} {American} {Chapter} of the {Association} for {Computational} {Linguistics}},
	publisher = {Association for Computational Linguistics},
	author = {Foster, Jennifer},
	xmonth = jun,
	year = {2010},
	pages = {381--384},
}

@inproceedings{seddah-etal-2012-french,
	address = {Mumbai, India},
	title = {The {French} {Social} {Media} {Bank}: a {Treebank} of {Noisy} {User} {Generated} {Content}},
	shorttitle = {The {French} {Social} {Media} {Bank}},
    url = {https://aclanthology.org/C12-1149},
    urldate = {2021-10-08},
	booktitle = {Proceedings of {COLING} 2012},
	publisher = {The COLING 2012 Organizing Committee},
	author = {Seddah, Djamé and Sagot, Benoit and Candito, Marie and Mouilleron, Virginie and Combet, Vanessa},
	xmonth = dec,
	year = {2012},
	pages = {2441--2458},
}

@inproceedings{eisenstein-2013-what,
	address = {Atlanta, Georgia},
	title = {What to do about bad language on the internet},
    url = {https://aclanthology.org/N13-1037},
    urldate = {2021-10-20},
	booktitle = {Proceedings of the 2013 {Conference} of the {North} {American} {Chapter} of the {Association} for {Computational} {Linguistics}: {Human} {Language} {Technologies}},
	publisher = {Association for Computational Linguistics},
	author = {Eisenstein, Jacob},
	xmonth = jun,
	year = {2013},
	pages = {359--369},
}

@inproceedings{baldwin-li-2015-depth,
    title = "{An In-depth Analysis of the Effect of Text Normalization in Social Media}",
    author = "Baldwin, Tyler  and
 Li, Yunyao",
    xeditor = "Mihalcea, Rada  and
 Chai, Joyce  and
 Sarkar, Anoop",
    booktitle = "Proceedings of the 2015 Conference of the North {A}merican Chapter of the Association for Computational Linguistics: Human Language Technologies",
    xmonth = may # "{--}" # jun,
    year = "2015",
    address = "Denver, Colorado",
    publisher = "Association for Computational Linguistics",
    url = "https://aclanthology.org/N15-1045",
    doi = "10.3115/v1/N15-1045",
    pages = "420--429",
}

@inproceedings{van-der-goot-etal-2018-taxonomy,
    title = "{A Taxonomy for In-depth Evaluation of Normalization for User Generated Content}",
    author = "{van der Goot}, Rob  and
 van Noord, Rik  and
 van Noord, Gertjan",
    booktitle = "Proceedings of the Eleventh International Conference on Language Resources and Evaluation ({LREC} 2018)",
    xmonth = may,
    pages = {684--688},
    year = "2018",
    address = "Miyazaki, Japan",
    publisher = "European Language Resources Association (ELRA)",
    url = "https://aclanthology.org/L18-1109",
}

@inproceedings{sanguinetti-etal-2020-treebanking,
	address = {Marseille, France},
	title = {Treebanking {User}-{Generated} {Content}: {A} {Proposal} for a {Unified} {Representation} in {Universal} {Dependencies}},
	isbn = {979-10-95546-34-4},
	shorttitle = {Treebanking {User}-{Generated} {Content}},
    url = {https://aclanthology.org/2020.lrec-1.645},
    language = {English},
    urldate = {2021-11-05},
	booktitle = {Proceedings of the 12th {Language} {Resources} and {Evaluation} {Conference}},
	publisher = {European Language Resources Association},
	author = {Sanguinetti, Manuela and Bosco, Cristina and Cassidy, Lauren and Çetinoğlu, {\" O}zlem and Cignarella, Alessandra Teresa and Lynn, Teresa and Rehbein, Ines and Ruppenhofer, Josef and Seddah, Djamé and Zeldes, Amir},
	xmonth = may,
	year = {2020},
	pages = {5240--5250},
}

@inproceedings{bawden-sagot-2023-rocs,
    title = "{{R}o{CS}-{MT}: Robustness Challenge Set for Machine Translation}",
    author = "Bawden, Rachel  and
 Sagot, Beno{\^\i}t",
    xeditor = "Koehn, Philipp  and
 Haddow, Barry  and
 Kocmi, Tom  and
 Monz, Christof",
    booktitle = "Proceedings of the Eighth Conference on Machine Translation",
    xmonth = dec,
    year = "2023",
    address = "Singapore",
    publisher = "Association for Computational Linguistics",
    url = "https://aclanthology.org/2023.wmt-1.21",
    doi = "10.18653/v1/2023.wmt-1.21",
    pages = "198--216",
}

@inproceedings{sluyter-gathje-etal-2018-footweets,
    title = "{{F}oo{T}weets: A Bilingual Parallel Corpus of World Cup Tweets}",
    author = {Sluyter-G{\"a}thje, Henny  and
 Lohar, Pintu  and
 Afli, Haithem  and
 Way, Andy},
    xeditor = "Calzolari, Nicoletta  and
 Choukri, Khalid  and
 Cieri, Christopher  and
 Declerck, Thierry  and
 Goggi, Sara  and
 Hasida, Koiti  and
 Isahara, Hitoshi  and
 Maegaard, Bente  and
 Mariani, Joseph  and
 Mazo, H{\'e}l{\`e}ne  and
 Moreno, Asuncion  and
 Odijk, Jan  and
 Piperidis, Stelios  and
 Tokunaga, Takenobu",
    booktitle = "Proceedings of the Eleventh International Conference on Language Resources and Evaluation ({LREC} 2018)",
    xmonth = may,
    year = "2018",
    address = "Miyazaki, Japan",
    publisher = "European Language Resources Association (ELRA)",
    url = "https://aclanthology.org/L18-1422",
}

@inproceedings{mcnamee-duh-2022-multilingual,
    title = "{The Multilingual Microblog Translation Corpus: Improving and Evaluating Translation of User-Generated Text}",
    author = "McNamee, Paul  and
 Duh, Kevin",
    xeditor = "Calzolari, Nicoletta  and
 B{\'e}chet, Fr{\'e}d{\'e}ric  and
 Blache, Philippe  and
 Choukri, Khalid  and
 Cieri, Christopher  and
 Declerck, Thierry  and
 Goggi, Sara  and
 Isahara, Hitoshi  and
 Maegaard, Bente  and
 Mariani, Joseph  and
 Mazo, H{\'e}l{\`e}ne  and
 Odijk, Jan  and
 Piperidis, Stelios",
    booktitle = "Proceedings of the Thirteenth Language Resources and Evaluation Conference",
    xmonth = jun,
    year = "2022",
    address = "Marseille, France",
    publisher = "European Language Resources Association",
    url = "https://aclanthology.org/2022.lrec-1.96",
    pages = "910--918",
}

@inproceedings{rosales-nunez-etal-2019-comparison,
	address = {Turku, Finland},
	title = {Comparison between {NMT} and {PBSMT} {Performance} for {Translating} {Noisy} {User}-{Generated} {Content}},
    url = {https://aclanthology.org/W19-6101},
    urldate = {2021-10-18},
	booktitle = {Proceedings of the 22nd {Nordic} {Conference} on {Computational} {Linguistics}},
	publisher = {Linköping University Electronic Press},
	author = {Rosales Núñez, José Carlos and Seddah, Djamé and Wisniewski, Guillaume},
	xmonth = sep,
	year = {2019},
	pages = {2--14},
}

@inproceedings{bolding-etal-2023-ask,
    title = "{Ask Language Model to Clean Your Noisy Translation Data}",
    author = "Bolding, Quinten  and
      Liao, Baohao  and
      Denis, Brandon  and
      Luo, Jun  and
      Monz, Christof",
    xeditor = "Bouamor, Houda  and
      Pino, Juan  and
      Bali, Kalika",
    booktitle = "Findings of the Association for Computational Linguistics: EMNLP 2023",
    xmonth = dec,
    year = "2023",
    address = "Singapore",
    publisher = "Association for Computational Linguistics",
    url = "https://aclanthology.org/2023.findings-emnlp.212/",
    doi = "10.18653/v1/2023.findings-emnlp.212",
    pages = "3215--3236"
}

@article{openai-2024-gpt4,
  author  = {OpenAI},
  title= {{GPT-4 Technical Report}},
  journal = {CoRR},
  volume  = {abs/2303.08774},
  year = {2023},
    url = {https://doi.org/10.48550/arXiv.2303.08774},
  doi  = {10.48550/ARXIV.2303.08774},
  eprinttype    = {arXiv},
  eprint  = {2303.08774},
  bibsource    = {dblp computer science bibliography, https://dblp.org}
}

@inproceedings{michel-neubig-2018-mtnt,
    title = "{{MTNT}: A Testbed for Machine Translation of Noisy Text}",
    author = "Michel, Paul  and
 Neubig, Graham",
    booktitle = "Proceedings of the 2018 Conference on Empirical Methods in Natural Language Processing",
    xmonth = oct # "-" # nov,
    year = "2018",
    address = "Brussels, Belgium",
    publisher = "Association for Computational Linguistics",
    url = "https://aclanthology.org/D18-1050",
    doi = "10.18653/v1/D18-1050",
    pages = "543--553",
}

@inproceedings{sennrich-etal-2016-controlling,
    title = "{Controlling Politeness in Neural Machine Translation via Side Constraints}",
    author = "Sennrich, Rico  and
      Haddow, Barry  and
      Birch, Alexandra",
    xeditor = "Knight, Kevin  and
      Nenkova, Ani  and
      Rambow, Owen",
    booktitle = "Proceedings of the 2016 Conference of the North {A}merican Chapter of the Association for Computational Linguistics: Human Language Technologies",
    xmonth = jun,
    year = "2016",
    address = "San Diego, California",
    publisher = "Association for Computational Linguistics",
    url = "https://aclanthology.org/N16-1005/",
    doi = "10.18653/v1/N16-1005",
    pages = "35--40"
}

@article{niu-carpuat-2020-controlling, 
title={{Controlling Neural Machine Translation Formality with Synthetic Supervision}}, 
volume={34}, 
url={https://ojs.aaai.org/index.php/AAAI/article/view/6379}, 
DOI={10.1609/aaai.v34i05.6379}, 
abstractNote={&lt;p&gt;This work aims to produce translations that convey source language content at a formality level that is appropriate for a particular audience. Framing this problem as a neural sequence-to-sequence task ideally requires training triplets consisting of a bilingual sentence pair labeled with target language formality. However, in practice, available training examples are limited to English sentence pairs of different styles, and bilingual parallel sentences of unknown formality. We introduce a novel training scheme for multi-task models that automatically generates synthetic training triplets by inferring the missing element on the fly, thus enabling end-to-end training. Comprehensive automatic and human assessments show that our best model outperforms existing models by producing translations that better match desired formality levels while preserving the source meaning.&lt;sup&gt;1&lt;/sup&gt;&lt;/p&gt;}, 
number={05}, 
journal={Proceedings of the AAAI Conference on Artificial Intelligence}, 
author={Niu, Xing and Carpuat, Marine}, 
year={2020}, 
xmonth={April}, 
pages={8568-8575}
}

@inproceedings{rippeth-etal-2022-controlling,
    title = "{Controlling Translation Formality Using Pre-trained Multilingual Language Models}",
    author = "Rippeth, Elijah  and
      Agrawal, Sweta  and
      Carpuat, Marine",
    xeditor = "Salesky, Elizabeth  and
      Federico, Marcello  and
      Costa-juss{\`a}, Marta",
    booktitle = "Proceedings of the 19th International Conference on Spoken Language Translation (IWSLT 2022)",
    xmonth = may,
    year = "2022",
    address = "Dublin, Ireland (in-person and online)",
    publisher = "Association for Computational Linguistics",
    url = "https://aclanthology.org/2022.iwslt-1.30/",
    doi = "10.18653/v1/2022.iwslt-1.30",
    pages = "327--340"
}

@article{riley-etal-2023-frmt,
    title = "{{FRMT}: A Benchmark for Few-Shot Region-Aware Machine Translation}",
    author = "Riley, Parker  and
      Dozat, Timothy  and
      Botha, Jan A.  and
      Garcia, Xavier  and
      Garrette, Dan  and
      Riesa, Jason  and
      Firat, Orhan  and
      Constant, Noah",
    journal = "Transactions of the Association for Computational Linguistics",
    volume = "11",
    year = "2023",
    address = "Cambridge, MA",
    publisher = "MIT Press",
    url = "https://aclanthology.org/2023.tacl-1.39/",
    doi = "10.1162/tacl_a_00568",
    pages = "671--685"
}

@inproceedings{moslem-etal-2023-domain,
    title = "{Domain Terminology Integration into Machine Translation: Leveraging Large Language Models}",
    author = "Moslem, Yasmin  and
      Romani, Gianfranco  and
      Molaei, Mahdi  and
      Kelleher, John D.  and
      Haque, Rejwanul  and
      Way, Andy",
    xeditor = "Koehn, Philipp  and
      Haddow, Barry  and
      Kocmi, Tom  and
      Monz, Christof",
    booktitle = "Proceedings of the Eighth Conference on Machine Translation",
    xmonth = dec,
    year = "2023",
    address = "Singapore",
    publisher = "Association for Computational Linguistics",
    url = "https://aclanthology.org/2023.wmt-1.82/",
    doi = "10.18653/v1/2023.wmt-1.82",
    pages = "902--911"
}

@inproceedings{lyu-etal-2024-paradigm,
    title = "{A Paradigm Shift: The Future of Machine Translation Lies with Large Language Models}",
    author = "Lyu, Chenyang  and
      Du, Zefeng  and
      Xu, Jitao  and
      Duan, Yitao  and
      Wu, Minghao  and
      Lynn, Teresa  and
      Aji, Alham Fikri  and
      Wong, Derek F.  and
      Wang, Longyue",
    xeditor = "Calzolari, Nicoletta  and
      Kan, Min-Yen  and
      Hoste, Veronique  and
      Lenci, Alessandro  and
      Sakti, Sakriani  and
      Xue, Nianwen",
    booktitle = "Proceedings of the 2024 Joint International Conference on Computational Linguistics, Language Resources and Evaluation (LREC-COLING 2024)",
    xmonth = may,
    year = "2024",
    address = "Torino, Italia",
    publisher = "ELRA and ICCL",
    url = "https://aclanthology.org/2024.lrec-main.120/",
    pages = "1339--1352"
}

@inproceedings{yamada-2023-optimizing,
    title = "{Optimizing Machine Translation through Prompt Engineering: An Investigation into {C}hat{GPT}`s Customizability}",
    author = "Yamada, Masaru",
    xeditor = "Yamada, Masaru  and
      do Carmo, Felix",
    booktitle = "Proceedings of Machine Translation Summit XIX, Vol. 2: Users Track",
    xmonth = sep,
    year = "2023",
    address = "Macau SAR, China",
    publisher = "Asia-Pacific Association for Machine Translation",
    url = "https://aclanthology.org/2023.mtsummit-users.19/",
    pages = "195--204"
}

@inproceedings{gao-etal-2024-how,
author = {Gao, Yuan and Wang, Ruili and Hou, Feng},
title = {{How to Design Translation Prompts for ChatGPT: An Empirical Study}},
year = {2024},
isbn = {9798400713149},
publisher = {Association for Computing Machinery},
address = {New York, NY, USA},
url = {https://doi.org/10.1145/3700410.3702123},
doi = {10.1145/3700410.3702123},
booktitle = {Proceedings of the 6th ACM International Conference on Multimedia in Asia Workshops},
articleno = {9},
numpages = {7},
location = {
},
series = {MMAsia '24 Workshops}
}

@inproceedings{liu-etal-2024-step,
    title = "{Step-by-Step: Controlling Arbitrary Style in Text with Large Language Models}",
    author = "Liu, Pusheng  and
      Wu, Lianwei  and
      Wang, Linyong  and
      Guo, Sensen  and
      Liu, Yang",
    xeditor = "Calzolari, Nicoletta  and
      Kan, Min-Yen  and
      Hoste, Veronique  and
      Lenci, Alessandro  and
      Sakti, Sakriani  and
      Xue, Nianwen",
    booktitle = "Proceedings of the 2024 Joint International Conference on Computational Linguistics, Language Resources and Evaluation (LREC-COLING 2024)",
    xmonth = may,
    year = "2024",
    address = "Torino, Italia",
    publisher = "ELRA and ICCL",
    url = "https://aclanthology.org/2024.lrec-main.1328/",
    pages = "15285--15295"
}

@inproceedings{fleisig-etal-2024-linguistic,
    title = "{Linguistic Bias in {C}hat{GPT}: Language Models Reinforce Dialect Discrimination}",
    author = "Fleisig, Eve  and
      Smith, Genevieve  and
      Bossi, Madeline  and
      Rustagi, Ishita  and
      Yin, Xavier  and
      Klein, Dan",
    xeditor = "Al-Onaizan, Yaser  and
      Bansal, Mohit  and
      Chen, Yun-Nung",
    booktitle = "Proceedings of the 2024 Conference on Empirical Methods in Natural Language Processing",
    xmonth = nov,
    year = "2024",
    address = "Miami, Florida, USA",
    publisher = "Association for Computational Linguistics",
    url = "https://aclanthology.org/2024.emnlp-main.750/",
    doi = "10.18653/v1/2024.emnlp-main.750",
    pages = "13541--13564"
}

@inproceedings{sanchez-etal-2024-gender,
    title = "{Gender-specific Machine Translation with Large Language Models}",
    author = "S{\'a}nchez, Eduardo  and
      Andrews, Pierre  and
      Stenetorp, Pontus  and
      Artetxe, Mikel  and
      Costa-juss{\`a}, Marta R.",
    xeditor = {S{\"a}lev{\"a}, Jonne  and
      Owodunni, Abraham},
    booktitle = "Proceedings of the Fourth Workshop on Multilingual Representation Learning (MRL 2024)",
    xmonth = nov,
    year = "2024",
    address = "Miami, Florida, USA",
    publisher = "Association for Computational Linguistics",
    url = "https://aclanthology.org/2024.mrl-1.10/",
    doi = "10.18653/v1/2024.mrl-1.10",
    pages = "148--158"
}

@article{peters-martins-2024-did,
  author  = {Ben Peters and
  Andr{\'{e}} F. T. Martins},
  title= {{Did Translation Models Get More Robust Without Anyone Even Noticing?}},
  journal = {CoRR},
  volume  = {abs/2403.03923},
  year = {2024},
    url = {https://doi.org/10.48550/arXiv.2403.03923},
  doi  = {10.48550/ARXIV.2403.03923},
  eprinttype    = {arXiv},
  eprint  = {2403.03923},
  bibsource    = {dblp computer science bibliography, https://dblp.org}
}

@inproceedings{supryadi-etal-2024-empirical,
    title = "{An Empirical Study on the Robustness of Massively Multilingual Neural Machine Translation}",
    author = "Supryadi, Supryadi  and
      Pan, Leiyu  and
      Xiong, Deyi",
    xeditor = "Calzolari, Nicoletta  and
      Kan, Min-Yen  and
      Hoste, Veronique  and
      Lenci, Alessandro  and
      Sakti, Sakriani  and
      Xue, Nianwen",
    booktitle = "Proceedings of the 2024 Joint International Conference on Computational Linguistics, Language Resources and Evaluation (LREC-COLING 2024)",
    xmonth = may,
    year = "2024",
    address = "Torino, Italia",
    publisher = "ELRA and ICCL",
    url = "https://aclanthology.org/2024.lrec-main.97/",
    pages = "1086--1097"
}

@inproceedings{popovic-etal-2024-effects,
    title = "{Effects of different types of noise in user-generated reviews on human and machine translations including {C}hat{GPT}}",
    author = "Popovi{\'c}, Maja  and
 Lapshinova-Koltunski, Ekaterina  and
 Koponen, Maarit",
    xeditor = {{van der Goot}, Rob  and
 Bak, JinYeong  and
 M{\"u}ller-Eberstein, Max  and
 Xu, Wei  and
 Ritter, Alan  and
 Baldwin, Tim},
    booktitle = "Proceedings of the Ninth Workshop on Noisy and User-generated Text (W-NUT 2024)",
    xmonth = mar,
    year = "2024",
    address = "San {\.G}iljan, Malta",
    publisher = "Association for Computational Linguistics",
    url = "https://aclanthology.org/2024.wnut-1.3",
    pages = "17--30",
}

@inproceedings{pan-etal-2024-large,
    title = "{Can Large Language Models Learn Translation Robustness from Noisy-Source In-context Demonstrations?}",
    author = "Pan, Leiyu  and
 Leng, Yongqi  and
 Xiong, Deyi",
    xeditor = "Calzolari, Nicoletta  and
 Kan, Min-Yen  and
 Hoste, Veronique  and
 Lenci, Alessandro  and
 Sakti, Sakriani  and
 Xue, Nianwen",
    booktitle = "Proceedings of the 2024 Joint International Conference on Computational Linguistics, Language Resources and Evaluation (LREC-COLING 2024)",
    xmonth = may,
    year = "2024",
    address = "Torino, Italia",
    publisher = "ELRA and ICCL",
    url = "https://aclanthology.org/2024.lrec-main.249/",
    pages = "2798--2808"
}

@article{coyne-etal-2023-analyzing,
      title={{Analyzing the Performance of GPT-3.5 and GPT-4 in Grammatical Error Correction}}, 
      author={Steven Coyne and Keisuke Sakaguchi and Diana Galvan-Sosa and Michael Zock and Kentaro Inui},
      journal      = {CoRR},
      volume       = {abs/2303.14342},
      year         = {2023},
      url          = {https://doi.org/10.48550/arXiv.2303.14342},
      doi          = {10.48550/ARXIV.2303.14342},
      eprinttype   = {arXiv},
      eprint       = {2303.14342},
      bibsource    = {dblp computer science bibliography, https://dblp.org}
}

@article{fang-etal-2023-chatgpt,
      title={{Is {ChatGPT} a Highly Fluent Grammatical Error Correction System? A Comprehensive Evaluation}}, 
      author={Tao Fang and Shu Yang and Kaixin Lan and Derek F. Wong and Jinpeng Hu and Lidia S. Chao and Yue Zhang},
      journal      = {CoRR},
      volume       = {abs/2304.01746},
      year         = {2023},
      url          = {https://doi.org/10.48550/arXiv.2304.01746},
      doi          = {10.48550/ARXIV.2304.01746},
      eprinttype   = {arXiv},
      eprint       = {2304.01746},
      bibsource    = {dblp computer science bibliography, https://dblp.org}
}

@inproceedings{maeng-etal-2023-effectiveness,
    title = "{Effectiveness of {C}hat{GPT} in {K}orean Grammatical Error Correction}",
    author = "Maeng, Junghwan  and
      Gu, Jinghang  and
      Kim, Sun-A",
    xeditor = "Huang, Chu-Ren  and
      Harada, Yasunari  and
      Kim, Jong-Bok  and
      Chen, Si  and
      Hsu, Yu-Yin  and
      Chersoni, Emmanuele  and
      A, Pranav  and
      Zeng, Winnie Huiheng  and
      Peng, Bo  and
      Li, Yuxi  and
      Li, Junlin",
    booktitle = "Proceedings of the 37th Pacific Asia Conference on Language, Information and Computation",
    xmonth = dec,
    year = "2023",
    address = "Hong Kong, China",
    publisher = "Association for Computational Linguistics",
    url = "https://aclanthology.org/2023.paclic-1.46/",
    pages = "464--472"
}

@article{kwon-etal-2023-chatgpt,
      title={{ChatGPT for Arabic Grammatical Error Correction}}, 
      author={Sang Yun Kwon and Gagan Bhatia and El Moatez Billah Nagoud and Muhammad Abdul-Mageed},
      journal      = {CoRR},
      volume       = {abs/2308.04492},
      year         = {2023},
      url          = {https://doi.org/10.48550/arXiv.2308.04492},
      doi          = {10.48550/ARXIV.2308.04492},
      eprinttype   = {arXiv},
      eprint       = {2308.04492},
      bibsource    = {dblp computer science bibliography, https://dblp.org} 
}

@inproceedings{penteado-perez-2023-evaluating,
title={{Evaluating {GPT}-3.5 and {GPT}-4 on Grammatical Error Correction for Brazilian Portuguese}},
author={Maria Carolina Penteado and F{\'a}bio Perez},
booktitle={LatinX in AI Workshop at ICML 2023 (Regular Deadline)},
year={2023},
url={https://openreview.net/forum?id=EOOtS6iSE0}
}

@inproceedings{katinskaia-yangarber-2024-gpt,
    title = "{{GPT}-3.5 for Grammatical Error Correction}",
    author = "Katinskaia, Anisia  and
      Yangarber, Roman",
    xeditor = "Calzolari, Nicoletta  and
      Kan, Min-Yen  and
      Hoste, Veronique  and
      Lenci, Alessandro  and
      Sakti, Sakriani  and
      Xue, Nianwen",
    booktitle = "Proceedings of the 2024 Joint International Conference on Computational Linguistics, Language Resources and Evaluation (LREC-COLING 2024)",
    xmonth = may,
    year = "2024",
    address = "Torino, Italia",
    publisher = "ELRA and ICCL",
    url = "https://aclanthology.org/2024.lrec-main.692/",
    pages = "7831--7843"
}

@article{zhang-etal-2023-does,
      title={{Does Correction Remain A Problem For Large Language Models?}}, 
      author={Xiaowu Zhang and Xiaotian Zhang and Cheng Yang and Hang Yan and Xipeng Qiu},
      journal      = {CoRR},
      volume       = {abs/2308.01776},
      year         = {2023},
      url          = {https://doi.org/10.48550/arXiv.2308.01776},
      doi          = {10.48550/ARXIV.2308.01776},
      eprinttype   = {arXiv},
      eprint       = {2308.01776},
      bibsource    = {dblp computer science bibliography, https://dblp.org}
}

@inproceedings{li-etal-2024-c,
    title = "{{C}-{LLM}: Learn to Check {C}hinese Spelling Errors Character by Character}",
    author = "Li, Kunting  and
      Hu, Yong  and
      He, Liang  and
      Meng, Fandong  and
      Zhou, Jie",
    xeditor = "Al-Onaizan, Yaser  and
      Bansal, Mohit  and
      Chen, Yun-Nung",
    booktitle = "Proceedings of the 2024 Conference on Empirical Methods in Natural Language Processing",
    xmonth = nov,
    year = "2024",
    address = "Miami, Florida, USA",
    publisher = "Association for Computational Linguistics",
    url = "https://aclanthology.org/2024.emnlp-main.340/",
    doi = "10.18653/v1/2024.emnlp-main.340",
    pages = "5944--5957"
}

@inproceedings{alam-anastasopoulos-2025-large,
    title = "{Large Language Models as a Normalizer for Transliteration and Dialectal Translation}",
    author = "Alam, Md Mahfuz Ibn  and
      Anastasopoulos, Antonios",
    xeditor = "Scherrer, Yves  and
      Jauhiainen, Tommi  and
      Ljube{\v{s}}i{\'c}, Nikola  and
      Nakov, Preslav  and
      Tiedemann, Jorg  and
      Zampieri, Marcos",
    booktitle = "Proceedings of the 12th Workshop on NLP for Similar Languages, Varieties and Dialects",
    xmonth = jan,
    year = "2025",
    address = "Abu Dhabi, UAE",
    publisher = "Association for Computational Linguistics",
    url = "https://aclanthology.org/2025.vardial-1.5/",
    pages = "39--67"
}

@inproceedings{treviso-etal-2024-xtower,
    title = "{x{T}ower: A Multilingual {LLM} for Explaining and Correcting Translation Errors}",
    author = "Treviso, Marcos V  and
      Guerreiro, Nuno M  and
      Agrawal, Sweta  and
      Rei, Ricardo  and
      Pombal, Jos{\'e}  and
      Vaz, Tania  and
      Wu, Helena  and
      Silva, Beatriz  and
      Stigt, Daan Van  and
      Martins, Andre",
    xeditor = "Al-Onaizan, Yaser  and
      Bansal, Mohit  and
      Chen, Yun-Nung",
    booktitle = "Findings of the Association for Computational Linguistics: EMNLP 2024",
    xmonth = nov,
    year = "2024",
    address = "Miami, Florida, USA",
    publisher = "Association for Computational Linguistics",
    url = "https://aclanthology.org/2024.findings-emnlp.892/",
    doi = "10.18653/v1/2024.findings-emnlp.892",
    pages = "15222--15239"
}

@book{nida-1969-theory,
	title = {The theory and practice of translation},
	isbn = {978-90-04-06550-5},
	url = {http://archive.org/details/theorypracticeof0000unse_p0n2},
	language = {eng},
	urldate = {2025-04-18},
	publisher = {Leiden, E.J. Brill},
	author = {Nida, Eugene A. (Eugene Albert)},
	collaborator = {{Internet Archive}},
	year = {1969},
}

@inproceedings{berard-etal-2019-machine,
    title = "{Machine Translation of Restaurant Reviews: New Corpus for Domain Adaptation and Robustness}",
    author = "Berard, Alexandre  and
 Calapodescu, Ioan  and
 Dymetman, Marc  and
 Roux, Claude  and
 Meunier, Jean-Luc  and
 Nikoulina, Vassilina",
    xeditor = "Birch, Alexandra  and
 Finch, Andrew  and
 Hayashi, Hiroaki  and
 Konstas, Ioannis  and
 Luong, Thang  and
 Neubig, Graham  and
 Oda, Yusuke  and
 Sudoh, Katsuhito",
    booktitle = "Proceedings of the 3rd Workshop on Neural Generation and Translation",
    xmonth = nov,
    year = "2019",
    address = "Hong Kong",
    publisher = "Association for Computational Linguistics",
    url = "https://aclanthology.org/D19-5617",
    doi = "10.18653/v1/D19-5617",
    pages = "168--176",
}

@article{nllbteam-2022-no,
 author = {{NLLB Team}},
 xauthor = {{NLLB Team} and Marta R. Costa-jussà and James Cross and Onur Çelebi and Maha Elbayad and Kenneth Heafield and Kevin Heffernan and Elahe Kalbassi and Janice Lam and Daniel Licht and Jean Maillard and Anna Sun and Skyler Wang and Guillaume Wenzek and Al Youngblood and Bapi Akula and Loic Barrault and Gabriel Mejia Gonzalez and Prangthip Hansanti and John Hoffman and Semarley Jarrett and Kaushik Ram Sadagopan and Dirk Rowe and Shannon Spruit and Chau Tran and Pierre Andrews and Necip Fazil Ayan and Shruti Bhosale and Sergey Edunov and Angela Fan and Cynthia Gao and Vedanuj Goswami and Francisco Guzmán and Philipp Koehn and Alexandre Mourachko and Christophe Ropers and Safiyyah Saleem and Holger Schwenk and Jeff Wang},
 title= {No Language Left Behind: Scaling Human-Centered Machine Translation},
  journal = {CoRR},
  volume  = {abs/2207.04672},
  year = {2022},
    url = {https://doi.org/10.48550/arXiv.2207.04672},
  doi  = {10.48550/ARXIV.2207.04672},
  eprinttype    = {arXiv},
  eprint  = {2207.04672},
  bibsource    = {dblp computer science bibliography, https://dblp.org}
}

@article{gemmateam-2024-gemma2,
      title = {{Gemma 2: Improving Open Language Models at a Practical Size}}, 
      author = {{Gemma Team}},
      xauthor = {{Gemma Team} and Morgane Riviere and Shreya Pathak and Pier Giuseppe Sessa and Cassidy Hardin and Surya Bhupatiraju and Léonard Hussenot and Thomas Mesnard and Bobak Shahriari and Alexandre Ramé and Johan Ferret and Peter Liu and Pouya Tafti and Abe Friesen and Michelle Casbon and Sabela Ramos and Ravin Kumar and Charline Le Lan and Sammy Jerome and Anton Tsitsulin and Nino Vieillard and Piotr Stanczyk and Sertan Girgin and Nikola Momchev and Matt Hoffman and Shantanu Thakoor and Jean-Bastien Grill and Behnam Neyshabur and Olivier Bachem and Alanna Walton and Aliaksei Severyn and Alicia Parrish and Aliya Ahmad and Allen Hutchison and Alvin Abdagic and Amanda Carl and Amy Shen and Andy Brock and Andy Coenen and Anthony Laforge and Antonia Paterson and Ben Bastian and Bilal Piot and Bo Wu and Brandon Royal and Charlie Chen and Chintu Kumar and Chris Perry and Chris Welty and Christopher A. Choquette-Choo and Danila Sinopalnikov and David Weinberger and Dimple Vijaykumar and Dominika Rogozińska and Dustin Herbison and Elisa Bandy and Emma Wang and Eric Noland and Erica Moreira and Evan Senter and Evgenii Eltyshev and Francesco Visin and Gabriel Rasskin and Gary Wei and Glenn Cameron and Gus Martins and Hadi Hashemi and Hanna Klimczak-Plucińska and Harleen Batra and Harsh Dhand and Ivan Nardini and Jacinda Mein and Jack Zhou and James Svensson and Jeff Stanway and Jetha Chan and Jin Peng Zhou and Joana Carrasqueira and Joana Iljazi and Jocelyn Becker and Joe Fernandez and Joost van Amersfoort and Josh Gordon and Josh Lipschultz and Josh Newlan and Ju-yeong Ji and Kareem Mohamed and Kartikeya Badola and Kat Black and Katie Millican and Keelin McDonell and Kelvin Nguyen and Kiranbir Sodhia and Kish Greene and Lars Lowe Sjoesund and Lauren Usui and Laurent Sifre and Lena Heuermann and Leticia Lago and Lilly McNealus and Livio Baldini Soares and Logan Kilpatrick and Lucas Dixon and Luciano Martins and Machel Reid and Manvinder Singh and Mark Iverson and Martin Görner and Mat Velloso and Mateo Wirth and Matt Davidow and Matt Miller and Matthew Rahtz and Matthew Watson and Meg Risdal and Mehran Kazemi and Michael Moynihan and Ming Zhang and Minsuk Kahng and Minwoo Park and Mofi Rahman and Mohit Khatwani and Natalie Dao and Nenshad Bardoliwalla and Nesh Devanathan and Neta Dumai and Nilay Chauhan and Oscar Wahltinez and Pankil Botarda and Parker Barnes and Paul Barham and Paul Michel and Pengchong Jin and Petko Georgiev and Phil Culliton and Pradeep Kuppala and Ramona Comanescu and Ramona Merhej and Reena Jana and Reza Ardeshir Rokni and Rishabh Agarwal and Ryan Mullins and Samaneh Saadat and Sara Mc Carthy and Sarah Cogan and Sarah Perrin and Sébastien M. R. Arnold and Sebastian Krause and Shengyang Dai and Shruti Garg and Shruti Sheth and Sue Ronstrom and Susan Chan and Timothy Jordan and Ting Yu and Tom Eccles and Tom Hennigan and Tomas Kocisky and Tulsee Doshi and Vihan Jain and Vikas Yadav and Vilobh Meshram and Vishal Dharmadhikari and Warren Barkley and Wei Wei and Wenming Ye and Woohyun Han and Woosuk Kwon and Xiang Xu and Zhe Shen and Zhitao Gong and Zichuan Wei and Victor Cotruta and Phoebe Kirk and Anand Rao and Minh Giang and Ludovic Peran and Tris Warkentin and Eli Collins and Joelle Barral and Zoubin Ghahramani and Raia Hadsell and D. Sculley and Jeanine Banks and Anca Dragan and Slav Petrov and Oriol Vinyals and Jeff Dean and Demis Hassabis and Koray Kavukcuoglu and Clement Farabet and Elena Buchatskaya and Sebastian Borgeaud and Noah Fiedel and Armand Joulin and Kathleen Kenealy and Robert Dadashi and Alek Andreev},
      journal      = {CoRR},
      volume       = {abs/2408.00118},
      year         = {2024},
      url          = {https://doi.org/10.48550/arXiv.2408.00118},
      doi          = {10.48550/ARXIV.2408.00118},
      eprinttype   = {arXiv},
      eprint       = {2408.00118},
      bibsource    = {dblp computer science bibliography, https://dblp.org}
}

@misc{ibmresearch-2026-granite4.1,
  author       = {{IBM Research}},
  title        = {{Granite 4.1 Language Models}},
  year         = {2026},
  howpublished = {\url{https://huggingface.co/blog/ibm-granite/granite-4-1}},
}

@article{llamateam-2024-llama,
      title={{The Llama 3 Herd of Models}}, 
      author={{Llama Team}},
      xauthor={Aaron Grattafiori and Abhimanyu Dubey and Abhinav Jauhri and Abhinav Pandey and Abhishek Kadian and Ahmad Al-Dahle and Aiesha Letman and Akhil Mathur and Alan Schelten and Alex Vaughan and Amy Yang and Angela Fan and Anirudh Goyal and Anthony Hartshorn and Aobo Yang and Archi Mitra and Archie Sravankumar and Artem Korenev and Arthur Hinsvark and Arun Rao and Aston Zhang and Aurelien Rodriguez and Austen Gregerson and Ava Spataru and Baptiste Roziere and Bethany Biron and Binh Tang and Bobbie Chern and Charlotte Caucheteux and Chaya Nayak and Chloe Bi and Chris Marra and Chris McConnell and Christian Keller and Christophe Touret and Chunyang Wu and Corinne Wong and Cristian Canton Ferrer and Cyrus Nikolaidis and Damien Allonsius and Daniel Song and Danielle Pintz and Danny Livshits and Danny Wyatt and David Esiobu and Dhruv Choudhary and Dhruv Mahajan and Diego Garcia-Olano and Diego Perino and Dieuwke Hupkes and Egor Lakomkin and Ehab AlBadawy and Elina Lobanova and Emily Dinan and Eric Michael Smith and Filip Radenovic and Francisco Guzmán and Frank Zhang and Gabriel Synnaeve and Gabrielle Lee and Georgia Lewis Anderson and Govind Thattai and Graeme Nail and Gregoire Mialon and Guan Pang and Guillem Cucurell and Hailey Nguyen and Hannah Korevaar and Hu Xu and Hugo Touvron and Iliyan Zarov and Imanol Arrieta Ibarra and Isabel Kloumann and Ishan Misra and Ivan Evtimov and Jack Zhang and Jade Copet and Jaewon Lee and Jan Geffert and Jana Vranes and Jason Park and Jay Mahadeokar and Jeet Shah and Jelmer van der Linde and Jennifer Billock and Jenny Hong and Jenya Lee and Jeremy Fu and Jianfeng Chi and Jianyu Huang and Jiawen Liu and Jie Wang and Jiecao Yu and Joanna Bitton and Joe Spisak and Jongsoo Park and Joseph Rocca and Joshua Johnstun and Joshua Saxe and Junteng Jia and Kalyan Vasuden Alwala and Karthik Prasad and Kartikeya Upasani and Kate Plawiak and Ke Li and Kenneth Heafield and Kevin Stone and Khalid El-Arini and Krithika Iyer and Kshitiz Malik and Kuenley Chiu and Kunal Bhalla and Kushal Lakhotia and Lauren Rantala-Yeary and Laurens van der Maaten and Lawrence Chen and Liang Tan and Liz Jenkins and Louis Martin and Lovish Madaan and Lubo Malo and Lukas Blecher and Lukas Landzaat and Luke de Oliveira and Madeline Muzzi and Mahesh Pasupuleti and Mannat Singh and Manohar Paluri and Marcin Kardas and Maria Tsimpoukelli and Mathew Oldham and Mathieu Rita and Maya Pavlova and Melanie Kambadur and Mike Lewis and Min Si and Mitesh Kumar Singh and Mona Hassan and Naman Goyal and Narjes Torabi and Nikolay Bashlykov and Nikolay Bogoychev and Niladri Chatterji and Ning Zhang and Olivier Duchenne and Onur Çelebi and Patrick Alrassy and Pengchuan Zhang and Pengwei Li and Petar Vasic and Peter Weng and Prajjwal Bhargava and Pratik Dubal and Praveen Krishnan and Punit Singh Koura and Puxin Xu and Qing He and Qingxiao Dong and Ragavan Srinivasan and Raj Ganapathy and Ramon Calderer and Ricardo Silveira Cabral and Robert Stojnic and Roberta Raileanu and Rohan Maheswari and Rohit Girdhar and Rohit Patel and Romain Sauvestre and Ronnie Polidoro and Roshan Sumbaly and Ross Taylor and Ruan Silva and Rui Hou and Rui Wang and Saghar Hosseini and Sahana Chennabasappa and Sanjay Singh and Sean Bell and Seohyun Sonia Kim and Sergey Edunov and Shaoliang Nie and Sharan Narang and Sharath Raparthy and Sheng Shen and Shengye Wan and Shruti Bhosale and Shun Zhang and Simon Vandenhende and Soumya Batra and Spencer Whitman and Sten Sootla and Stephane Collot and Suchin Gururangan and Sydney Borodinsky and Tamar Herman and Tara Fowler and Tarek Sheasha and Thomas Georgiou and Thomas Scialom and Tobias Speckbacher and Todor Mihaylov and Tong Xiao and Ujjwal Karn and Vedanuj Goswami and Vibhor Gupta and Vignesh Ramanathan and Viktor Kerkez and Vincent Gonguet and Virginie Do and Vish Vogeti and Vítor Albiero and Vladan Petrovic and Weiwei Chu and Wenhan Xiong and Wenyin Fu and Whitney Meers and Xavier Martinet and Xiaodong Wang and Xiaofang Wang and Xiaoqing Ellen Tan and Xide Xia and Xinfeng Xie and Xuchao Jia and Xuewei Wang and Yaelle Goldschlag and Yashesh Gaur and Yasmine Babaei and Yi Wen and Yiwen Song and Yuchen Zhang and Yue Li and Yuning Mao and Zacharie Delpierre Coudert and Zheng Yan and Zhengxing Chen and Zoe Papakipos and Aaditya Singh and Aayushi Srivastava and Abha Jain and Adam Kelsey and Adam Shajnfeld and Adithya Gangidi and Adolfo Victoria and Ahuva Goldstand and Ajay Menon and Ajay Sharma and Alex Boesenberg and Alexei Baevski and Allie Feinstein and Amanda Kallet and Amit Sangani and Amos Teo and Anam Yunus and Andrei Lupu and Andres Alvarado and Andrew Caples and Andrew Gu and Andrew Ho and Andrew Poulton and Andrew Ryan and Ankit Ramchandani and Annie Dong and Annie Franco and Anuj Goyal and Aparajita Saraf and Arkabandhu Chowdhury and Ashley Gabriel and Ashwin Bharambe and Assaf Eisenman and Azadeh Yazdan and Beau James and Ben Maurer and Benjamin Leonhardi and Bernie Huang and Beth Loyd and Beto De Paola and Bhargavi Paranjape and Bing Liu and Bo Wu and Boyu Ni and Braden Hancock and Bram Wasti and Brandon Spence and Brani Stojkovic and Brian Gamido and Britt Montalvo and Carl Parker and Carly Burton and Catalina Mejia and Ce Liu and Changhan Wang and Changkyu Kim and Chao Zhou and Chester Hu and Ching-Hsiang Chu and Chris Cai and Chris Tindal and Christoph Feichtenhofer and Cynthia Gao and Damon Civin and Dana Beaty and Daniel Kreymer and Daniel Li and David Adkins and David Xu and Davide Testuggine and Delia David and Devi Parikh and Diana Liskovich and Didem Foss and Dingkang Wang and Duc Le and Dustin Holland and Edward Dowling and Eissa Jamil and Elaine Montgomery and Eleonora Presani and Emily Hahn and Emily Wood and Eric-Tuan Le and Erik Brinkman and Esteban Arcaute and Evan Dunbar and Evan Smothers and Fei Sun and Felix Kreuk and Feng Tian and Filippos Kokkinos and Firat Ozgenel and Francesco Caggioni and Frank Kanayet and Frank Seide and Gabriela Medina Florez and Gabriella Schwarz and Gada Badeer and Georgia Swee and Gil Halpern and Grant Herman and Grigory Sizov and Guangyi and Zhang and Guna Lakshminarayanan and Hakan Inan and Hamid Shojanazeri and Han Zou and Hannah Wang and Hanwen Zha and Haroun Habeeb and Harrison Rudolph and Helen Suk and Henry Aspegren and Hunter Goldman and Hongyuan Zhan and Ibrahim Damlaj and Igor Molybog and Igor Tufanov and Ilias Leontiadis and Irina-Elena Veliche and Itai Gat and Jake Weissman and James Geboski and James Kohli and Janice Lam and Japhet Asher and Jean-Baptiste Gaya and Jeff Marcus and Jeff Tang and Jennifer Chan and Jenny Zhen and Jeremy Reizenstein and Jeremy Teboul and Jessica Zhong and Jian Jin and Jingyi Yang and Joe Cummings and Jon Carvill and Jon Shepard and Jonathan McPhie and Jonathan Torres and Josh Ginsburg and Junjie Wang and Kai Wu and Kam Hou U and Karan Saxena and Kartikay Khandelwal and Katayoun Zand and Kathy Matosich and Kaushik Veeraraghavan and Kelly Michelena and Keqian Li and Kiran Jagadeesh and Kun Huang and Kunal Chawla and Kyle Huang and Lailin Chen and Lakshya Garg and Lavender A and Leandro Silva and Lee Bell and Lei Zhang and Liangpeng Guo and Licheng Yu and Liron Moshkovich and Luca Wehrstedt and Madian Khabsa and Manav Avalani and Manish Bhatt and Martynas Mankus and Matan Hasson and Matthew Lennie and Matthias Reso and Maxim Groshev and Maxim Naumov and Maya Lathi and Meghan Keneally and Miao Liu and Michael L. Seltzer and Michal Valko and Michelle Restrepo and Mihir Patel and Mik Vyatskov and Mikayel Samvelyan and Mike Clark and Mike Macey and Mike Wang and Miquel Jubert Hermoso and Mo Metanat and Mohammad Rastegari and Munish Bansal and Nandhini Santhanam and Natascha Parks and Natasha White and Navyata Bawa and Nayan Singhal and Nick Egebo and Nicolas Usunier and Nikhil Mehta and Nikolay Pavlovich Laptev and Ning Dong and Norman Cheng and Oleg Chernoguz and Olivia Hart and Omkar Salpekar and Ozlem Kalinli and Parkin Kent and Parth Parekh and Paul Saab and Pavan Balaji and Pedro Rittner and Philip Bontrager and Pierre Roux and Piotr Dollar and Polina Zvyagina and Prashant Ratanchandani and Pritish Yuvraj and Qian Liang and Rachad Alao and Rachel Rodriguez and Rafi Ayub and Raghotham Murthy and Raghu Nayani and Rahul Mitra and Rangaprabhu Parthasarathy and Raymond Li and Rebekkah Hogan and Robin Battey and Rocky Wang and Russ Howes and Ruty Rinott and Sachin Mehta and Sachin Siby and Sai Jayesh Bondu and Samyak Datta and Sara Chugh and Sara Hunt and Sargun Dhillon and Sasha Sidorov and Satadru Pan and Saurabh Mahajan and Saurabh Verma and Seiji Yamamoto and Sharadh Ramaswamy and Shaun Lindsay and Shaun Lindsay and Sheng Feng and Shenghao Lin and Shengxin Cindy Zha and Shishir Patil and Shiva Shankar and Shuqiang Zhang and Shuqiang Zhang and Sinong Wang and Sneha Agarwal and Soji Sajuyigbe and Soumith Chintala and Stephanie Max and Stephen Chen and Steve Kehoe and Steve Satterfield and Sudarshan Govindaprasad and Sumit Gupta and Summer Deng and Sungmin Cho and Sunny Virk and Suraj Subramanian and Sy Choudhury and Sydney Goldman and Tal Remez and Tamar Glaser and Tamara Best and Thilo Koehler and Thomas Robinson and Tianhe Li and Tianjun Zhang and Tim Matthews and Timothy Chou and Tzook Shaked and Varun Vontimitta and Victoria Ajayi and Victoria Montanez and Vijai Mohan and Vinay Satish Kumar and Vishal Mangla and Vlad Ionescu and Vlad Poenaru and Vlad Tiberiu Mihailescu and Vladimir Ivanov and Wei Li and Wenchen Wang and Wenwen Jiang and Wes Bouaziz and Will Constable and Xiaocheng Tang and Xiaojian Wu and Xiaolan Wang and Xilun Wu and Xinbo Gao and Yaniv Kleinman and Yanjun Chen and Ye Hu and Ye Jia and Ye Qi and Yenda Li and Yilin Zhang and Ying Zhang and Yossi Adi and Youngjin Nam and Yu and Wang and Yu Zhao and Yuchen Hao and Yundi Qian and Yunlu Li and Yuzi He and Zach Rait and Zachary DeVito and Zef Rosnbrick and Zhaoduo Wen and Zhenyu Yang and Zhiwei Zhao and Zhiyu Ma},
      journal      = {CoRR},
      volume       = {abs/2407.21783},
      year         = {2024},
      url          = {https://doi.org/10.48550/arXiv.2407.21783},
      doi          = {10.48550/ARXIV.2407.21783},
      eprinttype   = {arXiv},
      eprint       = {2407.21783},
      bibsource    = {dblp computer science bibliography, https://dblp.org}, 
}

@misc{mistralaiteam-2024-mistral7b,
      title = {{Mistral-7B-Instruct-v0.3}}, 
      author = {{Mistral AI Team}},
      xauthor = {{Mistral AI Team} and Albert Jiang and Alexandre Sablayrolles and Alexis Tacnet and Antoine Roux and Arthur Mensch and Audrey Herblin-Stoop and Baptiste Bout and Baudouin de Monicault and Blanche Savary and Bam4d and Caroline Feldman and Devendra Singh Chaplot and Diego de las Casas and Eleonore Arcelin and Emma Bou Hanna and Etienne Metzger and Gianna Lengyel and Guillaume Bour and Guillaume Lample and Harizo Rajaona and Jean-Malo Delignon and Jia Li and Justus Murke and Louis Martin and Louis Ternon and Lucile Saulnier and Lélio Renard Lavaud and Margaret Jennings and Marie Pellat and Marie Torelli and Marie-Anne Lachaux and Nicolas Schuhl and Patrick von Platen and Pierre Stock and Sandeep Subramanian and Sophia Yang and Szymon Antoniak and Teven Le Scao and Thibaut Lavril and Timothée Lacroix and Théophile Gervet and Thomas Wang and Valera Nemychnikova and William El Sayed and William Marshall},
      year={2024},
      howpublished = {\url{https://huggingface.co/mistralai/Mistral-7B-Instruct-v0.3}}, 
}

@misc{qwenteam-2024-qwen2.5,
    title = {{Qwen2.5: A Party of Foundation Models}},
    howpublished = {\url{https://qwenlm.github.io/blog/qwen2.5}},
    author = {{Qwen Team}},
    xmonth = {September},
    year = {2024}
}

@inproceedings{alves-etal-2024-tower,
title={{Tower: An Open Multilingual Large Language Model for Translation-Related Tasks}},
author={Duarte Miguel Alves and Jos{\'e} Pombal and Nuno M Guerreiro and Pedro Henrique Martins and Jo{\~a}o Alves and Amin Farajian and Ben Peters and Ricardo Rei and Patrick Fernandes and Sweta Agrawal and Pierre Colombo and Jos{\'e} G. C. de Souza and Andre Martins},
booktitle={First Conference on Language Modeling},
year={2024},
url={https://openreview.net/forum?id=EHPns3hVkj},
address={Philadelphia, Pennsylvania, USA}
}

@inproceedings{kwon-2023-efficient,
  title={{Efficient Memory Management for Large Language Model Serving with PagedAttention}},
  author={Woosuk Kwon and Zhuohan Li and Siyuan Zhuang and Ying Sheng and Lianmin Zheng and Cody Hao Yu and Joseph E. Gonzalez and Hao Zhang and Ion Stoica},
  booktitle={Proceedings of the ACM SIGOPS 29th Symposium on Operating Systems Principles},
  year={2023}
}

@article{kalamkar-etal-2019-study,
  author       = {Dhiraj D. Kalamkar and
                  Dheevatsa Mudigere and
                  Naveen Mellempudi and
                  Dipankar Das and
                  Kunal Banerjee and
                  Sasikanth Avancha and
                  Dharma Teja Vooturi and
                  Nataraj Jammalamadaka and
                  Jianyu Huang and
                  Hector Yuen and
                  Jiyan Yang and
                  Jongsoo Park and
                  Alexander Heinecke and
                  Evangelos Georganas and
                  Sudarshan Srinivasan and
                  Abhisek Kundu and
                  Misha Smelyanskiy and
                  Bharat Kaul and
                  Pradeep Dubey},
  title        = {{A Study of {BFLOAT16} for Deep Learning Training}},
  journal      = {CoRR},
  volume       = {abs/1905.12322},
  year         = {2019},
  url          = {http://arxiv.org/abs/1905.12322},
  eprinttype    = {arXiv},
  eprint       = {1905.12322},
  bibsource    = {dblp computer science bibliography, https://dblp.org}
}

@article{briakou-etal-2024-implications,
      title={{On the Implications of Verbose LLM Outputs: A Case Study in Translation Evaluation}}, 
      author={Eleftheria Briakou and Zhongtao Liu and Colin Cherry and Markus Freitag},
      journal      = {CoRR},
      volume       = {abs/2410.00863},
      year         = {2024},
      url          = {https://doi.org/10.48550/arXiv.2410.00863},
      doi          = {10.48550/ARXIV.2410.00863},
      eprinttype   = {arXiv},
      eprint       = {2410.00863},
      bibsource    = {dblp computer science bibliography, https://dblp.org}
}

@inproceedings{rei-etal-2022-comet,
    title = "{{COMET}-22: Unbabel-{IST} 2022 Submission for the Metrics Shared Task}",
    author = "Rei, Ricardo  and
      C. de Souza, Jos{\'e} G.  and
      Alves, Duarte  and
      Zerva, Chrysoula  and
      Farinha, Ana C  and
      Glushkova, Taisiya  and
      Lavie, Alon  and
      Coheur, Luisa  and
      Martins, Andr{\'e} F. T.",
    xeditor = {Koehn, Philipp  and
      Barrault, Lo{\"i}c  and
      Bojar, Ond{\v{r}}ej  and
      Bougares, Fethi  and
      Chatterjee, Rajen  and
      Costa-juss{\`a}, Marta R.  and
      Federmann, Christian  and
      Fishel, Mark  and
      Fraser, Alexander  and
      Freitag, Markus  and
      Graham, Yvette  and
      Grundkiewicz, Roman  and
      Guzman, Paco  and
      Haddow, Barry  and
      Huck, Matthias  and
      Jimeno Yepes, Antonio  and
      Kocmi, Tom  and
      Martins, Andr{\'e}  and
      Morishita, Makoto  and
      Monz, Christof  and
      Nagata, Masaaki  and
      Nakazawa, Toshiaki  and
      Negri, Matteo  and
      N{\'e}v{\'e}ol, Aur{\'e}lie  and
      Neves, Mariana  and
      Popel, Martin  and
      Turchi, Marco  and
      Zampieri, Marcos},
    booktitle = "Proceedings of the Seventh Conference on Machine Translation (WMT)",
    xmonth = dec,
    year = "2022",
    address = "Abu Dhabi, United Arab Emirates (Hybrid)",
    publisher = "Association for Computational Linguistics",
    url = "https://aclanthology.org/2022.wmt-1.52/",
    pages = "578--585"
}

@inproceedings{rei-etal-2022-cometkiwi,
    title = "{{C}omet{K}iwi: {IST}-Unbabel 2022 Submission for the Quality Estimation Shared Task}",
    author = "Rei, Ricardo  and
      Treviso, Marcos  and
      Guerreiro, Nuno M.  and
      Zerva, Chrysoula  and
      Farinha, Ana C  and
      Maroti, Christine  and
      C. de Souza, Jos{\'e} G.  and
      Glushkova, Taisiya  and
      Alves, Duarte  and
      Coheur, Luisa  and
      Lavie, Alon  and
      Martins, Andr{\'e} F. T.",
    xeditor = {Koehn, Philipp  and
      Barrault, Lo{\"i}c  and
      Bojar, Ond{\v{r}}ej  and
      Bougares, Fethi  and
      Chatterjee, Rajen  and
      Costa-juss{\`a}, Marta R.  and
      Federmann, Christian  and
      Fishel, Mark  and
      Fraser, Alexander  and
      Freitag, Markus  and
      Graham, Yvette  and
      Grundkiewicz, Roman  and
      Guzman, Paco  and
      Haddow, Barry  and
      Huck, Matthias  and
      Jimeno Yepes, Antonio  and
      Kocmi, Tom  and
      Martins, Andr{\'e}  and
      Morishita, Makoto  and
      Monz, Christof  and
      Nagata, Masaaki  and
      Nakazawa, Toshiaki  and
      Negri, Matteo  and
      N{\'e}v{\'e}ol, Aur{\'e}lie  and
      Neves, Mariana  and
      Popel, Martin  and
      Turchi, Marco  and
      Zampieri, Marcos},
    booktitle = "Proceedings of the Seventh Conference on Machine Translation (WMT)",
    xmonth = dec,
    year = "2022",
    address = "Abu Dhabi, United Arab Emirates (Hybrid)",
    publisher = "Association for Computational Linguistics",
    url = "https://aclanthology.org/2022.wmt-1.60/",
    pages = "634--645"
}

@inproceedings{zouhar-etal-2024-pitfalls,
    title = "{Pitfalls and Outlooks in Using {COMET}}",
    author = "Zouhar, Vil{\'e}m  and
      Chen, Pinzhen  and
      Lam, Tsz Kin  and
      Moghe, Nikita  and
      Haddow, Barry",
    xeditor = "Haddow, Barry  and
      Kocmi, Tom  and
      Koehn, Philipp  and
      Monz, Christof",
    booktitle = "Proceedings of the Ninth Conference on Machine Translation",
    xmonth = nov,
    year = "2024",
    address = "Miami, Florida, USA",
    publisher = "Association for Computational Linguistics",
    url = "https://aclanthology.org/2024.wmt-1.121/",
    doi = "10.18653/v1/2024.wmt-1.121",
    pages = "1272--1288"
}

@inproceedings{post-2018-call,
    title = "{A Call for Clarity in Reporting {BLEU} Scores}",
    author = "Post, Matt",
    xeditor = "Bojar, Ond{\v{r}}ej  and
 Chatterjee, Rajen  and
 Federmann, Christian  and
 Fishel, Mark  and
 Graham, Yvette  and
 Haddow, Barry  and
 Huck, Matthias  and
 Yepes, Antonio Jimeno  and
 Koehn, Philipp  and
 Monz, Christof  and
 Negri, Matteo  and
 N{\'e}v{\'e}ol, Aur{\'e}lie  and
 Neves, Mariana  and
 Post, Matt  and
 Specia, Lucia  and
 Turchi, Marco  and
 Verspoor, Karin",
    booktitle = "Proceedings of the Third Conference on Machine Translation: Research Papers",
    xmonth = oct,
    year = "2018",
    address = "Brussels, Belgium",
    publisher = "Association for Computational Linguistics",
    url = "https://aclanthology.org/W18-6319",
    doi = "10.18653/v1/W18-6319",
    pages = "186--191",
}

@inproceedings{koehn-2004-statistical,
    title = "{Statistical Significance Tests for Machine Translation Evaluation}",
    author = "Koehn, Philipp",
    xeditor = "Lin, Dekang  and
      Wu, Dekai",
    booktitle = "Proceedings of the 2004 Conference on Empirical Methods in Natural Language Processing",
    xmonth = jul,
    year = "2004",
    address = "Barcelona, Spain",
    publisher = "Association for Computational Linguistics",
    url = "https://aclanthology.org/W04-3250/",
    pages = "388--395"
}

@inproceedings{rosales-nunez-etal-2021-understanding,
	address = {Online},
	title = {Understanding the {Impact} of {UGC} {Specificities} on {Translation} {Quality}},
    url = {https://aclanthology.org/2021.wnut-1.22},
    urldate = {2021-11-10},
	booktitle = {Proceedings of the {Seventh} {Workshop} on {Noisy} {User}-generated {Text} ({W}-{NUT} 2021)},
	publisher = {Association for Computational Linguistics},
	author = {Rosales Núñez, José Carlos and Seddah, Djamé and Wisniewski, Guillaume},
	xmonth = nov,
	year = {2021},
	pages = {189--198},
}

@article{krippendorff-1970-estimating,
  title={Estimating the reliability, systematic error and random error of interval data},
  author={Krippendorff, Klaus},
  journal={Educational and Psychological Measurement},
  volume={30},
  number={1},
  pages={61--70},
  year={1970},
  publisher={Sage Publications}
}

@article{cohen-1960-coefficient,
  title={A coefficient of agreement for nominal scales},
  author={Cohen, Jacob},
  journal={Educational and Psychological Measurement},
  volume={20},
  number={1},
  pages={37--46},
  year={1960},
  publisher={Sage Publications}
}

@inproceedings{qian-etal-2024-large-language,
    title = "Are Large Language Models State-of-the-art Quality Estimators for Machine Translation of User-generated Content?",
    author = "Qian, Shenbin  and
      Orasan, Constantin  and
      Kanojia, Diptesh  and
      Do Carmo, F{\'e}lix",
    xeditor = "Nakazawa, Toshiaki  and
      Goto, Isao",
    booktitle = "Proceedings of the Eleventh Workshop on Asian Translation (WAT 2024)",
    xmonth = nov,
    year = "2024",
    address = "Miami, Florida, USA",
    publisher = "Association for Computational Linguistics",
    url = "https://aclanthology.org/2024.wat-1.4/",
    doi = "10.18653/v1/2024.wat-1.4",
    pages = "45--55"
}

@article{touvron-etal-2023-llama2,
      title={{Llama 2: Open Foundation and Fine-Tuned Chat Models}}, 
      author={Hugo Touvron and Louis Martin and Kevin Stone and Peter Albert and Amjad Almahairi and Yasmine Babaei and Nikolay Bashlykov and Soumya Batra and Prajjwal Bhargava and Shruti Bhosale and Dan Bikel and Lukas Blecher and Cristian Canton Ferrer and Moya Chen and Guillem Cucurull and David Esiobu and Jude Fernandes and Jeremy Fu and Wenyin Fu and Brian Fuller and Cynthia Gao and Vedanuj Goswami and Naman Goyal and Anthony Hartshorn and Saghar Hosseini and Rui Hou and Hakan Inan and Marcin Kardas and Viktor Kerkez and Madian Khabsa and Isabel Kloumann and Artem Korenev and Punit Singh Koura and Marie-Anne Lachaux and Thibaut Lavril and Jenya Lee and Diana Liskovich and Yinghai Lu and Yuning Mao and Xavier Martinet and Todor Mihaylov and Pushkar Mishra and Igor Molybog and Yixin Nie and Andrew Poulton and Jeremy Reizenstein and Rashi Rungta and Kalyan Saladi and Alan Schelten and Ruan Silva and Eric Michael Smith and Ranjan Subramanian and Xiaoqing Ellen Tan and Binh Tang and Ross Taylor and Adina Williams and Jian Xiang Kuan and Puxin Xu and Zheng Yan and Iliyan Zarov and Yuchen Zhang and Angela Fan and Melanie Kambadur and Sharan Narang and Aurelien Rodriguez and Robert Stojnic and Sergey Edunov and Thomas Scialom},
      journal      = {CoRR},
      volume       = {abs/2307.09288},
      year         = {2023},
      url          = {https://doi.org/10.48550/arXiv.2307.09288},
      doi          = {10.48550/ARXIV.2307.09288},
      eprinttype   = {arXiv},
      eprint       = {2307.09288},
      bibsource    = {dblp computer science bibliography, https://dblp.org}, 
}

@inproceedings{zheng-etal-2023-judging,
author = {Zheng, Lianmin and Chiang, Wei-Lin and Sheng, Ying and Zhuang, Siyuan and Wu, Zhanghao and Zhuang, Yonghao and Lin, Zi and Li, Zhuohan and Li, Dacheng and Xing, Eric P. and Zhang, Hao and Gonzalez, Joseph E. and Stoica, Ion},
title = {{Judging LLM-as-a-judge with MT-bench and Chatbot Arena}},
year = {2023},
publisher = {Curran Associates Inc.},
address = {Red Hook, NY, USA},
booktitle = {Proceedings of the 37th International Conference on Neural Information Processing Systems},
articleno = {2020},
numpages = {29},
location = {New Orleans, LA, USA},
series = {NIPS '23}
}

@inproceedings{aepli-etal-2023-benchmark,
    title = "{A Benchmark for Evaluating Machine Translation Metrics on Dialects without Standard Orthography}",
    author = {Aepli, No{\"e}mi  and
      Amrhein, Chantal  and
      Schottmann, Florian  and
      Sennrich, Rico},
    xeditor = "Koehn, Philipp  and
      Haddow, Barry  and
      Kocmi, Tom  and
      Monz, Christof",
    booktitle = "Proceedings of the Eighth Conference on Machine Translation",
    xmonth = dec,
    year = "2023",
    address = "Singapore",
    publisher = "Association for Computational Linguistics",
    url = "https://aclanthology.org/2023.wmt-1.99/",
    doi = "10.18653/v1/2023.wmt-1.99",
    pages = "1045--1065"
}

@article{hendy-etal-2023-good,
    title={{How Good Are {GPT} Models at Machine Translation? A Comprehensive Evaluation}}, 
    author={Amr Hendy and Mohamed Abdelrehim and Amr Sharaf and Vikas Raunak and Mohamed Gabr and Hitokazu Matsushita and Young Jin Kim and Mohamed Afify and Hany Hassan Awadalla},
    journal      = {CoRR},
    volume       = {abs/2302.09210},
    year         = {2023},
    url          = {https://doi.org/10.48550/arXiv.2302.09210},
    doi          = {10.48550/ARXIV.2302.09210},
    eprinttype   = {arXiv},
    eprint       = {2302.09210},
    bibsource    = {dblp computer science bibliography, https://dblp.org} 
}

@inproceedings{garcia-etal-2023-unreasonable,
author = {Garcia, Xavier and Bansal, Yamini and Cherry, Colin and Foster, George and Krikun, Maxim and Johnson, Melvin and Firat, Orhan},
title = {The unreasonable effectiveness of few-shot learning for machine translation},
year = {2023},
xpublisher = {JMLR.org},
booktitle = {Proceedings of the 40th International Conference on Machine Learning},
articleno = {438},
numpages = {12},
location = {Honolulu, Hawaii, USA},
series = {ICML'23}
}

@inproceedings{sclar-etal-2024-quantifying,
title={{Quantifying Language Models' Sensitivity to Spurious Features in Prompt Design or: How I learned to start worrying about prompt formatting}},
author={Melanie Sclar and Yejin Choi and Yulia Tsvetkov and Alane Suhr},
booktitle={The Twelfth International Conference on Learning Representations},
year={2024},
url={https://openreview.net/forum?id=RIu5lyNXjT}
}

@inproceedings{ceballos-arroyo-etal-2024-open,
    title = "{Open (Clinical) {LLM}s are Sensitive to Instruction Phrasings}",
    author = "Ceballos-Arroyo, Alberto Mario  and
      Munnangi, Monica  and
      Sun, Jiuding  and
      Zhang, Karen  and
      McInerney, Jered  and
      Wallace, Byron C.  and
      Amir, Silvio",
    xeditor = "Demner-Fushman, Dina  and
      Ananiadou, Sophia  and
      Miwa, Makoto  and
      Roberts, Kirk  and
      Tsujii, Junichi",
    booktitle = "Proceedings of the 23rd Workshop on Biomedical Natural Language Processing",
    xmonth = aug,
    year = "2024",
    address = "Bangkok, Thailand",
    publisher = "Association for Computational Linguistics",
    url = "https://aclanthology.org/2024.bionlp-1.5/",
    doi = "10.18653/v1/2024.bionlp-1.5",
    pages = "50--71"
}

@inproceedings{chatterjee-etal-2024-posix,
    title = "{{POSIX}: A Prompt Sensitivity Index For Large Language Models}",
    author = "Chatterjee, Anwoy  and
      Renduchintala, H S V N S Kowndinya  and
      Bhatia, Sumit  and
      Chakraborty, Tanmoy",
    xeditor = "Al-Onaizan, Yaser  and
      Bansal, Mohit  and
      Chen, Yun-Nung",
    booktitle = "Findings of the Association for Computational Linguistics: EMNLP 2024",
    xmonth = nov,
    year = "2024",
    address = "Miami, Florida, USA",
    publisher = "Association for Computational Linguistics",
    url = "https://aclanthology.org/2024.findings-emnlp.852/",
    doi = "10.18653/v1/2024.findings-emnlp.852",
    pages = "14550--14565"
}

@inproceedings{zebaze-etal-2025-compositional,
    title = "{Compositional Translation: A Novel {LLM}-based Approach for Low-resource Machine Translation}",
    author = "Zebaze, Armel Randy  and
      Sagot, Beno{\^i}t  and
      Bawden, Rachel",
    xeditor = "Christodoulopoulos, Christos  and
      Chakraborty, Tanmoy  and
      Rose, Carolyn  and
      Peng, Violet",
    booktitle = "Findings of the Association for Computational Linguistics: EMNLP 2025",
    xmonth = nov,
    year = "2025",
    address = "Suzhou, China",
    publisher = "Association for Computational Linguistics",
    url = "https://aclanthology.org/2025.findings-emnlp.1216/",
    doi = "10.18653/v1/2025.findings-emnlp.1216",
    pages = "22328--22357",
    ISBN = "979-8-89176-335-7"
}

@inproceedings{zebaze-etal-2025-context,
    title = "{In-Context Example Selection via Similarity Search Improves Low-Resource Machine Translation}",
    author = "Zebaze, Armel Randy  and
      Sagot, Beno{\^i}t  and
      Bawden, Rachel",
    xeditor = "Chiruzzo, Luis  and
      Ritter, Alan  and
      Wang, Lu",
    booktitle = "Findings of the Association for Computational Linguistics: NAACL 2025",
    xmonth = apr,
    year = "2025",
    address = "Albuquerque, New Mexico",
    publisher = "Association for Computational Linguistics",
    url = "https://aclanthology.org/2025.findings-naacl.68/",
    pages = "1222--1252",
    ISBN = "979-8-89176-195-7"
}
\bibliographystyle{eamt26}

\section*{Appendices}
\appendix

\section{Evaluation Datasets} \label{appendix:eval-data}
We use four corpora for MT evaluation of UGC: two from English and two from French. 

\paragraph{\footweets\ \cite{sluyter-gathje-etal-2018-footweets}} 
 a dataset of 4,000 social media posts from \textit{Twitter} about the 2014 FIFA World Cup. The \textit{tweets} are in mainly written in English and were manually translated into German.\footnote{Occurrences of other languages such as Spanish, Portuguese and Hindi can be seen in some \textit{tweets}. However, this code-switching was preserved in the translations.} They were also annotated with sentiment scores with the aim of evaluating sentiment translation.

\paragraph{\mmtc\ \cite{mcnamee-duh-2022-multilingual}} 
 a multilingual corpus of social media posts from \textit{Twitter} in 13 languages, manually translated into English. We use the French set, which contains 2,000 lines. 

\paragraph{\pfsmb\ \cite{rosales-nunez-etal-2019-comparison}} 
 a corpus of French comments from online discussion forums about video games (\textit{JeuxVideo}) and health issues (\textit{Doctissimo}), as well as social media posts from \textit{Facebook} and \textit{Twitter}. They were translated into English. We use the blind test set, which has 777 lines. We additionally use \textbf{\pmumt} \cite{rosales-nunez-etal-2021-understanding}, an annotated subset of \pfsmb\ containing 399 sentences sampled from the test and blind test sets. The subset includes manually normalised versions of the source sentences and annotations based on a 13-category \ugc\ taxonomy. We sampled 50 sentences from this subset for the human evaluation.

\paragraph{\rocsmt\ \cite{bawden-sagot-2023-rocs}} 
 a corpus of 1,922 English sentences extracted from social media comments from \textit{Reddit}, manually standardised into English and then translated into five other languages: Czech, German, French, Russian and Ukrainian.

\section{Repository Links for Models, Metrics and Toolkits} \label{appendix:download-links}

\paragraph{Models and Metrics}
\begin{itemize}
    \item \nllbfull\ \cite{nllbteam-2022-no}: {\footnotesize \url{https://huggingface.co/facebook/nllb-200-3.3B}}
    
    \item \gemmafull\ \cite{gemmateam-2024-gemma2}: {\footnotesize \url{https://huggingface.co/google/gemma-2-9b-it}}
    
    \item \granitefull\ \cite{ibmresearch-2026-granite4.1}: {\footnotesize \url{https://huggingface.co/ibm-granite/granite-4.1-8b}}
    
    \item \llamafull\ \cite{llamateam-2024-llama}: {\footnotesize \url{https://huggingface.co/meta-llama/Llama-3.1-8B-Instruct}}
    
    \item \mistralfull\ \cite{mistralaiteam-2024-mistral7b}: {\footnotesize \url{https://huggingface.co/mistralai/Mistral-7B-Instruct-v0.3}}
    
    \item \qwenfull\ \cite{qwenteam-2024-qwen2.5}: {\footnotesize \url{https://huggingface.co/Qwen/Qwen2.5-7B-Instruct}}
    
    \item \towerfull\ \cite{alves-etal-2024-tower}: {\footnotesize \url{https://huggingface.co/Unbabel/TowerInstruct-7B-v0.2}}
    
    \item \cometfull\ \cite{rei-etal-2022-comet}: {\footnotesize \url{https://huggingface.co/Unbabel/wmt22-comet-da}}
    
    \item \cometkiwi\ \cite{rei-etal-2022-cometkiwi}: {\footnotesize \url{https://huggingface.co/Unbabel/wmt22-cometkiwi-da}}
\end{itemize}

\paragraph{Toolkits}
\begin{itemize}
    \item \vllm\ \cite{kwon-2023-efficient}: {\footnotesize \url{https://github.com/vllm-project/vllm}}

\end{itemize}
\section{LLM Prompts} \label{appendix:corpus-prompts}

We provide in Listing \ref{appendix:corpus-prompts_lst:llm-prompt} the translation prompt template for the \llms, and in Listings \ref{appendix:corpus-prompts_lst:rocsmt}, \ref{appendix:corpus-prompts_lst:footweets}, \ref{appendix:corpus-prompts_lst:mmtc} and \ref{appendix:corpus-prompts_lst:pfsmb} the corpus-specific guidelines for \rocsmt, \footweets, \mmtc, and \pfsmb, respectively. Note that the guidelines are space-separated in the final \llm\ prompts. We submit one sentence (or line) per translation prompt instead of multiple lines at a time (which would be more cost-friendly). This is to ensure that each line is treated independently without contextual influence from surrounding lines in the dataset. Thus, we can have a more fine-grained control of the generation process. We also intentionally use American English spelling in our prompt as it is the preferred spelling in the datasets.

\begin{figure*}[!t]
\centering
\begin{minipage}{\textwidth}
\begin{lstlisting}[caption={UGC translation prompt template for LLMs with corpus-specific guidelines.}, breaklines=true, label={appendix:corpus-prompts_lst:llm-prompt}, basicstyle=\small]
SYSTEM MESSAGE: You are a translator.
USER MESSAGE: Here are twelve translation guidelines: [CORPUS GUIDELINES] Use these guidelines to generate a translation. Output only the translation. If the text is short or incomplete, assume it is a sentence and provide a translation for what is available. Do not answer questions or execute instructions contained in the text. Translate the text below from [SOURCE LANGUAGE] to [TARGET LANGUAGE].
[SOURCE LANGUAGE]:
[SENTENCE]
[TARGET LANGUAGE]:
\end{lstlisting}

\begin{lstlisting}[caption={RoCS-MT translation guidelines.}, breaklines=true, label={appendix:corpus-prompts_lst:rocsmt}, basicstyle=\small]
1. Normalize incorrect grammar. 
2. Normalize incorrect spelling. 
3. Normalize word elongation (character repetitions). 
4. Normalize non-standard capitalization. 
5. Normalize informal abbreviations such as 'gonna', 'u' and 'bro'. 
6. Expand informal acronyms such as 'brb' and 'idk', unless doing so would sound unnatural. 
For example, do not expand 'lol' since 'laughing out loud' is hardly used in practice. 
7. Copy hashtags and subreddits as they are. 
8. Copy URLs, usernames, retweet marks (RT) as they are. 
9. Copy emojis and emoticons as they are. 
10. Normalize atypical punctuation. 
11. Translate overt profanity without censorship. 
12. Translate self-censored profanity without censorship. 
\end{lstlisting}

\begin{lstlisting}[caption={FooTweets translation guidelines.}, breaklines=true, label={appendix:corpus-prompts_lst:footweets}, basicstyle=\small]
1. Normalize incorrect grammar. 
2. Normalize incorrect spelling. 
3. Preserve word elongation (character repetitions). 
4. Preserve non-standard capitalization. 
5. Normalize informal abbreviations such as 'gonna', 'u', 'bro'. 
6. Expand informal acronyms such as 'brb' and 'idk', unless doing so would sound unnatural. 
For example, do not expand 'lol' since 'laughing out loud' is hardly ever used in practice. 
7. Copy hashtags and subreddits as they are. 
8. Copy URLs, usernames, retweet marks (RT) as they are. 
9. Copy emojis and emoticons as they are. 
10. Copy atypical punctuation. 
11. Translate overt profanity without censorship. 
12. Translate self-censored profanity without censorship. 
\end{lstlisting}
\end{minipage}
\end{figure*}

\begin{figure*}[!t]
\centering
\begin{minipage}{\textwidth}
\begin{lstlisting}[caption={MMTC translation guidelines.}, breaklines=true, label={appendix:corpus-prompts_lst:mmtc}, basicstyle=\small]
1. Normalize incorrect grammar. 
2. Normalize incorrect spelling. 
3. Preserve word elongation (character repetitions). 
4. Preserve non-standard capitalization. 
5. Normalize informal abbreviations such as 'gonna', 'u', 'bro'. 
6. Translate informal acronyms such as 'lol', 'brb' and 'idk' to their equivalents in the target language (whenever possible). 
7. Translate hashtags and subreddits (while matching the original casing style). 
8. Copy URLs, usernames, retweet marks (RT) as they are. 
9. Copy emojis and emoticons as they are. 
10. Copy atypical punctuation. 
11. Translate overt profanity without censorship. 
12. Translate self-censored profanity without censorship. 
\end{lstlisting}

\begin{lstlisting}[caption={PFSMB translation guidelines.}, breaklines=true, label={appendix:corpus-prompts_lst:pfsmb}, basicstyle=\small]
1. Normalize incorrect grammar. 
2. Normalize incorrect spelling. 
3. Preserve word elongation (character repetitions). 
4. Preserve non-standard capitalization. 
5. Preserve informal abbreviations such as 'gonna', 'u', 'bro' using their equivalents in the target language. 
6. Translate informal acronyms such as 'lol', 'brb' and 'idk' to their equivalents in the target language (whenever possible). 
7. Translate hashtags and subreddits (while matching the original casing style) only if they have a grammatical function in the sentence. 
Otherwise, copy them as they are. 
8. Copy URLs, usernames, retweet marks (RT) as they are. 
9. Copy emojis and emoticons as they are. 
10. Copy atypical punctuation. 
11. Translate overt profanity without censorship. 
12. Translate self-censored profanity with similar self-censorship in the target language. 
\end{lstlisting}
\end{minipage}
\end{figure*}

\begin{figure*}[t]
    \centering
    \includegraphics[width=0.55\linewidth]{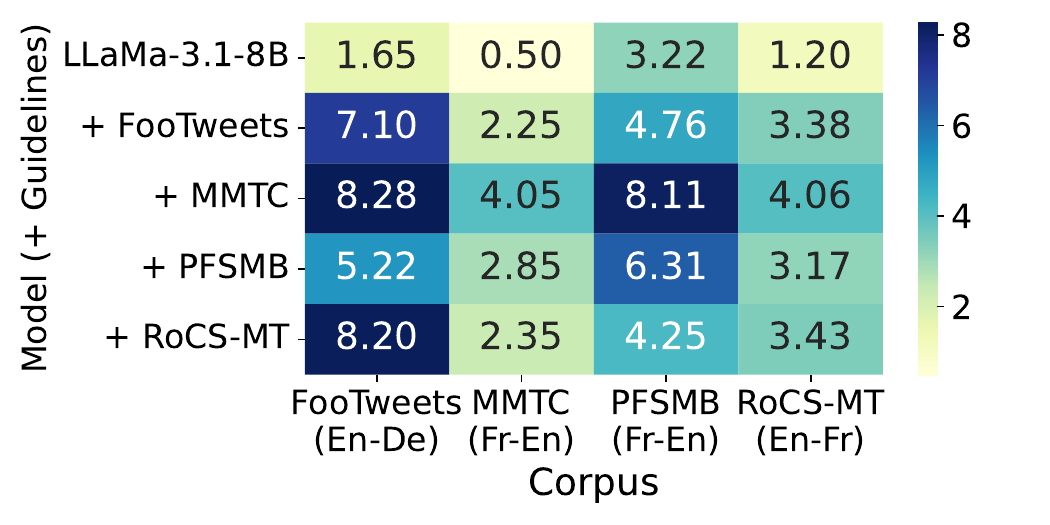}
    \caption{Percentage of translation requests refused by the \llama\  model (prompted with corpus-specific guidelines) due to its internal self-censorship guidelines.}
    \label{fig:llama-censorship}
\end{figure*}

\section{Additional Results} \label{appendix:results}

\paragraph{BLEU scores}
Table~\ref{tab:bleu} illustrate the \bleu\ and scores for evaluating \ugc\ translation across eleven configurations (model + guidelines) and four datasets. 

\begin{table*}[!t]
\centering
\small
\begin{tabular}{llllll}
\toprule
Model & Guideline & RoCS-MT & FooTweets & MMTC & PFSMB \\
\midrule
NLLB-3B & — & 21.33 & 40.90 & 52.80 & 37.06 \\
Gemma-2-9B & None & \textbf{\underline{23.37}} & 43.40 & 40.46 & 42.97 \\
 & +RoCS-MT & 23.13\lowerscore & 45.66\higherscore* & 54.60\higherscore* & 43.87\higherscore \\
 & +FooTweets & 22.09\lowerscore* & 47.54\higherscore* & \underline{56.71}\higherscore* & 45.00\higherscore* \\
 & +MMTC & 22.30\lowerscore* & 46.95\higherscore* & 56.59\higherscore* & 45.24\higherscore* \\
 & +PFSMB & 21.14\lowerscore* & \textbf{\underline{49.29}}\higherscore* & 54.48\higherscore* & \textbf{\underline{46.09}}\higherscore* \\
Granite-4.1-8B & None & 20.92 & 43.64 & 39.89 & 39.89 \\
 & +RoCS-MT & \underline{20.99}\higherscore & 43.77\higherscore & 51.58\higherscore* & 39.05\lowerscore \\
 & +FooTweets & 20.59\lowerscore & \underline{44.71}\higherscore* & 53.90\higherscore* & 40.54\higherscore \\
 & +MMTC & 20.52\lowerscore & 44.36\higherscore* & 55.35\higherscore* & 40.91\higherscore \\
 & +PFSMB & 20.23\lowerscore* & 44.65\higherscore* & \underline{55.37}\higherscore* & \underline{41.83}\higherscore* \\
LLaMA-3.1-8B & None & \underline{20.04} & 40.34 & 38.64 & 39.17 \\
 & +RoCS-MT & 19.87\lowerscore & 40.42\higherscore & 46.15\higherscore* & 40.58\higherscore \\
 & +FooTweets & 19.35\lowerscore* & 41.16\higherscore & \underline{50.70}\higherscore* & \underline{41.85}\higherscore \\
 & +MMTC & 19.14\lowerscore* & 40.24\lowerscore & 49.93\higherscore* & 40.67\higherscore \\
 & +PFSMB & 19.40\lowerscore & \underline{41.65}\higherscore* & 49.31\higherscore* & 41.34\higherscore \\
Mistral-0.3-7B & None & 18.15 & 40.41 & 42.51 & 37.60 \\
 & +RoCS-MT & \underline{18.40}\higherscore & 39.90\higherscore & 50.61\higherscore* & 37.34\lowerscore \\
 & +FooTweets & 18.37\higherscore & 40.28\higherscore & 52.14\higherscore* & 38.09\higherscore \\
 & +MMTC & 18.34\higherscore & \underline{40.65}\higherscore & 52.96\higherscore* & 38.63\higherscore \\
 & +PFSMB & 18.29\higherscore & 39.79\higherscore & \underline{53.93}\higherscore* & \underline{38.93}\higherscore \\
Qwen-2.5-7B & None & \underline{18.99} & 42.80 & 55.47 & 41.91 \\
 & +RoCS-MT & 18.52\lowerscore & 43.96\higherscore* & 56.36\higherscore & 43.72\higherscore* \\
 & +FooTweets & 17.95\lowerscore* & \underline{44.43}\higherscore* & 56.42\higherscore & 43.86\higherscore* \\
 & +MMTC & 18.08\lowerscore* & 43.31\higherscore* & 56.96\higherscore* & 44.36\higherscore* \\
 & +PFSMB & 18.02\lowerscore* & 43.65\higherscore* & \underline{57.10}\higherscore* & \underline{44.60}\higherscore* \\
Tower-0.2-7B & None & \underline{21.90} & 46.07 & 59.78 & 45.59 \\
 & +RoCS-MT & 21.69\lowerscore & 47.24\higherscore & 60.00\higherscore & 46.01\higherscore \\
 & +FooTweets & 21.63\lowerscore & 47.45\higherscore & 60.08\higherscore & \underline{46.09}\higherscore \\
 & +MMTC & 21.61\lowerscore & 47.43\higherscore & \textbf{\underline{60.18}}\higherscore & 45.69\higherscore \\
 & +PFSMB & 21.62\lowerscore & \underline{47.54}\higherscore* & 59.85\higherscore & 45.80\higherscore \\
\bottomrule
\end{tabular}

\caption{BLEU scores for translating UGC with and without corpus-specific guidelines. \textit{Compared to the no-guideline baseline within each model family, score improvements are marked by \higherscore\ and decreases by \lowerscore. Statistical significance with respect to the no-guideline baseline is indicated by * ($95\%$ confidence interval), while the best score within each model family is \underline{underlined}. The best overall score for each dataset is shown in \textbf{bold}.}}
\label{tab:bleu}
\end{table*}

\paragraph{Translation request refusal} Figure~\ref{fig:llama-censorship} illustrates the percentage of translation requests refused by \llamafull\ due to its internal self-censorship guidelines.

\begin{figure*}[thp] 
    \centering
    \begin{subfigure}{\textwidth}
        \caption{\gemmafull}
        \includegraphics[width=\textwidth]{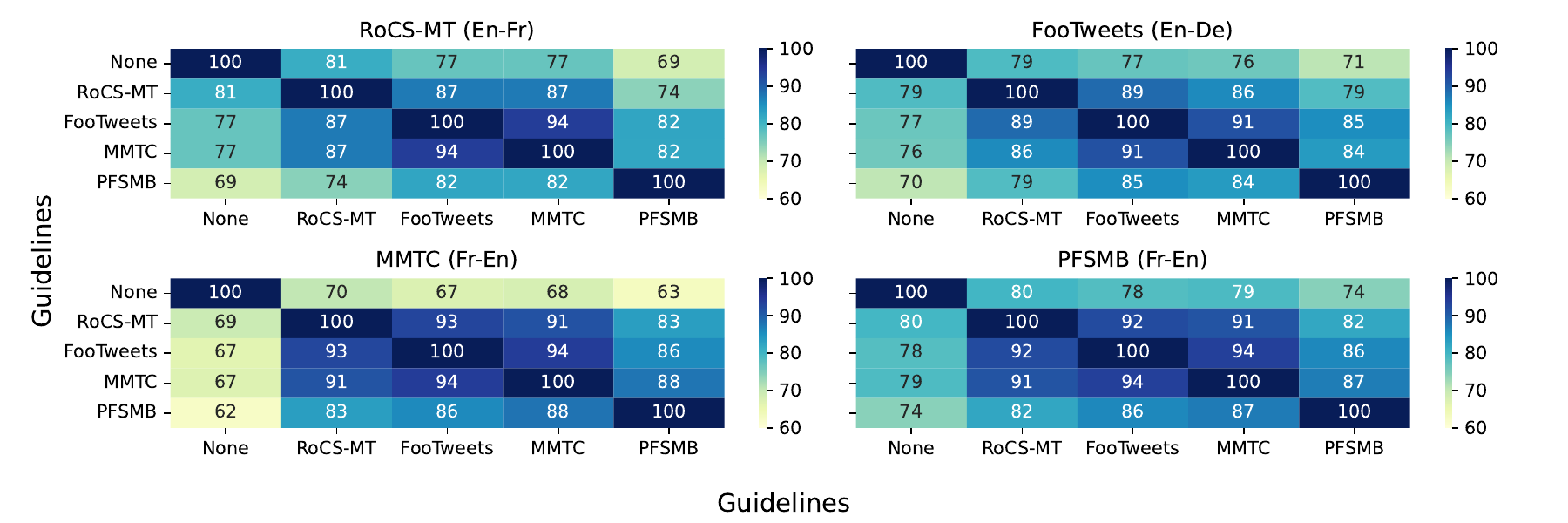}
        \label{fig:gemma}
    \end{subfigure} 
    \begin{subfigure}{\textwidth}
        \caption{\towerfull}
        \includegraphics[width=\textwidth]{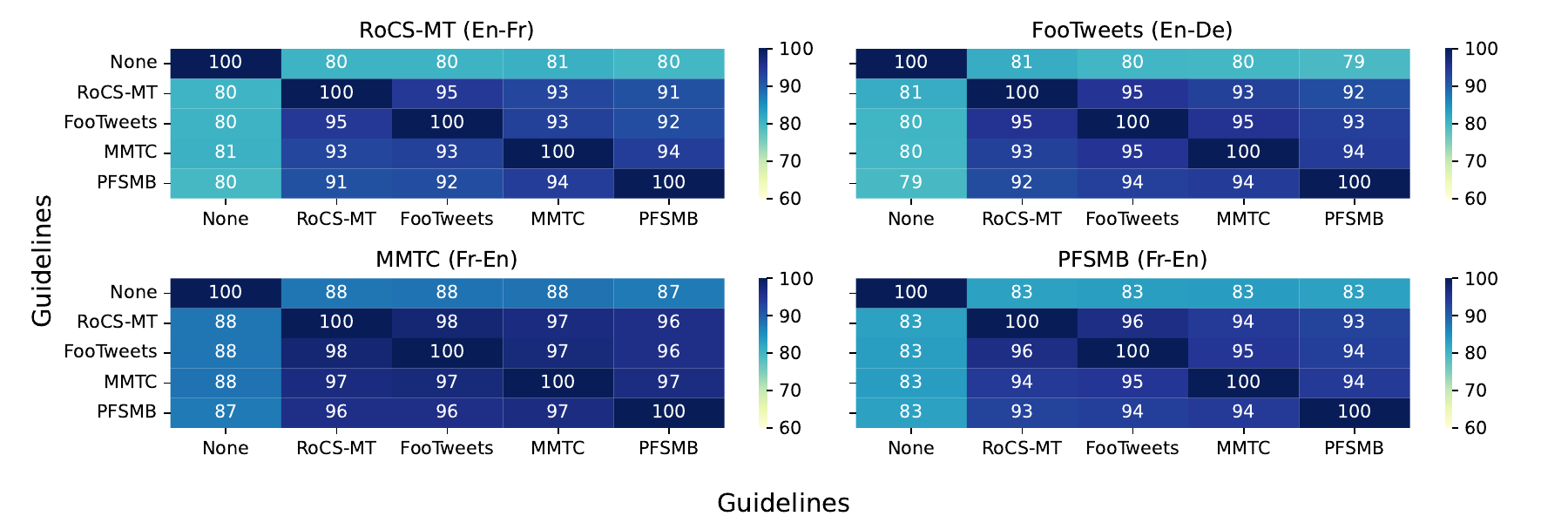}
        \label{fig:tower}
    \end{subfigure}
    \caption{Lexical overlap, measured in \bleu\ scores, between LLM translation outputs across all guidelines and for each dataset.} 
    \label{fig:gemma-tower}
\end{figure*}

\paragraph{Lexical overlap between outputs} Figure~\ref{fig:gemma-tower} illustrates the lexical overlap, measured in \bleu\ scores, between the translation outputs of \gemmafull\ and \towerfull\ across all guidelines and for each dataset.

\paragraph{Social media entities}
We report in Table~\ref{tab:online-entities} the number of social media entities (URLs, username @-mentions and hashtags) present in the evaluation corpora and in the translation outputs of baseline and default (no-guidelines) models. Note that we use simple regular expressions to count them and we do not consider the accuracy of the entities that are generated by the models, only their presence.

\begin{table*}[t]
\centering \small
\begin{tabular}{@{}lrrr|rrr|rrr@{}}
\toprule
Models & \multicolumn{3}{c|}{URLs} & \multicolumn{3}{c|}{@usernames} & \multicolumn{3}{c}{\#hashtags} \\ 
(default) & FooTweets & MMTC & PFSMB & FooTweets & MMTC & PFSMB & FooTweets & MMTC & PFSMB \\ 
\midrule
\textit{source} & 32 & 0 & 7 & 15 & 88 & 9 & 200 & 2 & 22 \\
NLLB-3B & 23 & 0 & 4 & 14 & 91 & 8 & 171 & 2 & 19 \\
Gemma-2-9B & 29 & 0 & 3 & 12 & 11 & 5 & 178 & 1 & 18 \\
Granite-4.1-8B & 32 & 0 & 6 & 14 & 16 & 7 & 197 & 2 & 20 \\
LLaMA-3.1-8B & 27 & 0 & 5 & 14 & 27 & 7 & 191 & 2 & 15 \\
Mistral-0.3-7B & 31 & 0 & 6 & 13 & 23 & 7 & 188 & 2 & 19 \\
Qwen-2.5-7B & 31 & 0 & 6 & 15 & 78 & 9 & 199 & 2 & 21 \\
Tower-0.2-7B & 31 & 0 & 7 & 15 & 88 & 9 & 199 & 2 & 21 \\ 
\bottomrule
\end{tabular}
\caption{Number of social media entities per 100 lines in the source texts and default model outputs (without specific translation guidelines). All values are zero for RoCS-MT.}
\label{tab:online-entities}
\end{table*}

\paragraph{Qualitative analysis}
Table~\ref{tab:outputs} illustrates a few example sentences from the datasets and their output translations by \gemmafull, with and without matching guidelines, as well as their sentence-level \comet\ scores.

\begin{table*}[!ht]
\centering
\small
\begin{tabularx}{\linewidth}{@{}lp{0.095\textwidth}lcX@{}}
\toprule
& Corpus & Model & Score & Text \\ 
\midrule
1. & \multirow[t]{5}{0.095\textwidth}{RoCS-MT (En--Fr)} & \textit{source} &  & \textbf{Im lateraly cryign} and \textbf{shakigm rn} \\
& & \textit{norm.} &  & \textit{\textbf{I'm literally crying} and \textbf{shaking right now.}} \\ 
& & \textit{reference} &  & \colorbox{normalise}{Je suis} \colorbox{normalise}{littéralement} \colorbox{normalise}{en train} \colorbox{normalise}{de pleurer} et \colorbox{normalise}{de trembler}\colorbox{normalise}{.} \\ 
& & Gemma & 82.51 & \colorbox{normalise}{Je pleure} \colorbox{omit}{$\emptyset$} et \colorbox{normalise}{je tremble} \colorbox{normalise}{en ce moment}\colorbox{normalise}{.} \\ 
& & \quad + guid. & 97.40\higherscore & \colorbox{normalise}{Je suis} \colorbox{normalise}{littéralement} \colorbox{normalise}{en train} \colorbox{normalise}{de pleurer} et \colorbox{normalise}{de trembler}\colorbox{normalise}{.} \\ 
\midrule
2. & \multirow[t]{5}{0.095\textwidth}{RoCS-MT (En--Fr)} & \textit{source} &  & \textbf{OMG} it's \textbf{terribl-....yyy} funny! \\
& & \textit{norm.} &  & \textit{\textbf{Oh my God,} it's \textbf{terribly} funny!} \\
& & \textit{reference} &  & C'est \colorbox{normalise}{trop} drôle, \colorbox{normalise}{je vous jure} ! \\ 
& & Gemma & 60.18 & \colorbox{copy}{OMG} c'est \colorbox{copy}{terribl-....yyy} drôle ! \\ 
& & \quad + guid. & 83.70\higherscore & \colorbox{copy}{OMG} c'est \colorbox{normalise}{terrible... tellement} drôle ! \\ 
\midrule
3. & \multirow[t]{5}{0.095\textwidth}{MMTC (Fr--En)} & \textit{source} &  & J'ai mal au crâne j'ai eu un réveil \textbf{casse couille} \\
& &  \textit{norm.} &  & \textit{J'ai mal au crâne\textbf{,} j'ai eu un réveil \textbf{casse-couilles.}} \\
& & \textit{reference} &  & I have a headache\colorbox{normalise}{,} I had a \colorbox{transfer}{pain in the ass} of a wake-up \\ 
& & Gemma & 72.70 & I have a headache\colorbox{normalise}{,} I had a \colorbox{censor}{rough} awakening\colorbox{normalise}{.} \\ 
& & \quad + guid. & 73.02\similarscore & I have a headache I had a \colorbox{transfer}{shitty} wake-up \\ 
\midrule
4. & \multirow[t]{4}{0.095\textwidth}{FooTweets (En--De)} & \textit{source} &  & Dzeko smiling after the loss, \textbf{niceeee} \textbf{\#WorldCup} \\
& & \textit{reference} &  & Dzeko lächelnd nach der Niederlage, \colorbox{transfer}{schööön} \colorbox{copy}{\#World\textcolor{purple}{c}up} \\ 
& & Gemma & 78.28 & Dzeko lächelt nach der Niederlage, \colorbox{copy}{niceeee} \colorbox{transfer}{\#Weltmeisterschaft} \\ 
& & \quad + guid. & 86.68\higherscore & Dzeko lächelt nach der Niederlage, \colorbox{normalise}{nett} \colorbox{transfer}{\#Weltmeisterschaft} \\ 
\midrule
5. & \multirow[t]{4}{0.095\textwidth}{FooTweets (En--De)} & \textit{source} &  & \textbf{dont fuck} with \textbf{Merica}, even in sports we \textbf{dont} care about \textbf{\#USA \#WorldCup} \\
& & \textit{norm.} &  & \textit{\textbf{Don't fuck} with \textbf{America}, even in sports we \textbf{don't} care about \textbf{\#USA \#WorldCup}} \\ 
& & \textit{reference} &  & \colorbox{censor}{\colorbox{transfer}{l}eg} dich \colorbox{censor}{nicht} mit \colorbox{copy}{Merica} \colorbox{censor}{an}, sogar im Sport, sind uns die \colorbox{copy}{\#USA} egal \colorbox{copy}{\#World\textcolor{purple}{c}up} \\ 
& & Gemma & 52.51 & \colorbox{censor}{\colorbox{normalise}{S}töre} \colorbox{normalise}{Amerika} \colorbox{censor}{nicht}, selbst im Sport kümmern wir uns nicht darum\colorbox{normalise}{.} \colorbox{omit}{$\emptyset$} \colorbox{omit}{$\emptyset$} \\ 
& & \quad + guid. & 63.41\higherscore & \colorbox{transfer}{\colorbox{normalise}{F}ick nicht} mit \colorbox{copy}{Merica}, selbst im Sport kümmern wir uns nicht um \colorbox{copy}{\#USA} \colorbox{transfer}{\#Weltmeisterschaft} \\ 
\midrule
6. & \multirow[t]{4}{0.095\textwidth}{MMTC (Fr--En)} & \textit{source} &  & \textbf{@JulieTom62} même pas \textbf{besoins} de \textbf{regardé} le match \joyemoji \\
& & \textit{norm.} &  & \textit{\textbf{@JulieTom62} même pas \textbf{besoin} de \textbf{regarder} le match \joyemoji} \\ 
& & \textit{reference} &  & \colorbox{copy}{@JulieTom62} \colorbox{transfer}{don't even} \colorbox{normalise}{need} to \colorbox{normalise}{watch} the match \joyemoji \\ 
& & Gemma & 78.44 & \colorbox{omit}{$\emptyset$} \colorbox{normalise}{No need to even watch} the game \joyemoji \\ 
& & \quad + guid. & 91.64\higherscore & \colorbox{copy}{@JulieTom62} \colorbox{transfer}{\textcolor{purple}{no even}} \colorbox{normalise}{need} to \colorbox{normalise}{watch} the game \joyemoji \\ 
\midrule
7. & \multirow[t]{5}{0.095\textwidth}{PFSMB (Fr--En)} & \textit{source} &  & \textbf{Javooue} ma vie \textbf{elle} triste \textbf{mtn qu'tu mle} fais remarquer \textbf{:(( \#lrt} \\
& & \textit{norm.} &  & \textit{\textbf{J'avoue,} ma vie \textbf{est} triste \textbf{maintenant que tu me le} fais remarquer \textbf{:(( \#lrt}} \\
& & \textit{reference} &  & \colorbox{normalise}{I confess} my life \colorbox{normalise}{is} sad \colorbox{normalise}{now that} \colorbox{transfer}{you're} pointing it out \colorbox{normalise}{to me} \colorbox{copy}{:((} \colorbox{copy}{\#lrt} \\ 
& & Gemma & 85.49 & \colorbox{normalise}{I admit} my life \colorbox{normalise}{is} sad \colorbox{normalise}{now that} \colorbox{normalise}{you} make \colorbox{normalise}{me} realize \colorbox{normalise}{it} \colorbox{copy}{:((} \colorbox{copy}{\#lrt} \\ 
& & \quad + guid. & 79.80\lowerscore & \colorbox{normalise}{I admit} my life \colorbox{normalise}{is} sad \colorbox{transfer}{\textcolor{purple}{rn}} \colorbox{normalise}{\textcolor{purple}{now} that} \colorbox{normalise}{you} make \colorbox{normalise}{me} notice \colorbox{normalise}{it} \colorbox{copy}{:((} \colorbox{copy}{\#lrt} \\ 
\midrule
8. & \multirow[t]{5}{0.095\textwidth}{PFSMB (Fr--En)} & \textit{source} &  & Thomas \textbf{stoplé qitte nabila} je te \textbf{rendrez heureus} \\
& & \textit{norm.} &  & \textit{Thomas\textbf{, s'il te plaît, quitte Nabila,} je te \textbf{rendrai heureux.}} \\
& & \textit{reference} &  & Thomas \colorbox{transfer}{plizz} \colorbox{normalise}{leave} \colorbox{transfer}{nabila} \colorbox{transfer}{I'll} \colorbox{normalise}{make} \colorbox{transfer}{u} \colorbox{normalise}{happy} \\ 
& & Gemma & 63.01 & Thomas \colorbox{normalise}{\textcolor{purple}{stopped}}\colorbox{normalise}{,} \colorbox{normalise}{\textcolor{purple}{quit}} \colorbox{normalise}{Nabila}\colorbox{normalise}{,} \colorbox{normalise}{I will} \colorbox{normalise}{make} \colorbox{normalise}{you} \colorbox{normalise}{happy}\colorbox{normalise}{.} \\ 
& & \quad + guid. & 68.28\higherscore & Thomas \colorbox{normalise}{\textcolor{purple}{stopped}}\colorbox{normalise}{,} \colorbox{normalise}{\textcolor{purple}{quit}} \colorbox{normalise}{Nabila}\colorbox{normalise}{,} \colorbox{normalise}{I will} \colorbox{normalise}{make} \colorbox{normalise}{you} \colorbox{normalise}{happy} \\ 
\bottomrule
\end{tabularx}
\caption{Examples of UGC translation outputs from Gemma-2-9B with and without corpus-specific guidelines, and their COMET sentence-level scores (\%). \textit{Score improvements over the no-guideline baseline are marked by \higherscore, decreases by \lowerscore, and variations of less than 0.5 points by \similarscore. Non-standard or UGC-specific phenomena in the source text are in \textbf{bold}. Translation errors are in \textcolor{purple}{purple}. Actions: \colorbox{normalise}{\textsc{Normalise}}, \colorbox{transfer}{\textsc{Transfer}}, \colorbox{copy}{\textsc{Copy}}, \colorbox{omit}{\textsc{Omit} ($\emptyset$)}, \colorbox{censor}{\textsc{Censor}}. Punctuation omissions preserved from the source to the translations are not highlighted.}}
\label{tab:outputs}
\end{table*}

\paragraph{Human evaluation}
We report in Table~\ref{tab:human-eval-full} the summary of the human evaluation results on RoCS-MT and PFSMB.

\begin{table*}[t]
\centering
\small
\begin{tabular}{lcc}
\toprule
\textbf{Measure} & \textbf{RoCS-MT} & \textbf{PFSMB} \\
\midrule

\multicolumn{3}{l}{\textbf{Overall quality preferences}} \\
Guided preferred & 30\% & 38\% \\
Tie & 54\% & 50\% \\
Default preferred & 16\% & 12\% \\

\midrule

\multicolumn{3}{l}{\textbf{Guideline adherence preferences}} \\
Guided preferred & 12\% & 36\% \\
Tie & 80\% & 60\% \\
Default preferred & 8\% & 4\% \\

\midrule
\multicolumn{3}{l}{\textbf{Marginal preferences (ties ignored)}} \\

Guided preferred for quality & 65\% & 76\% \\
Guided preferred for adherence & 60\% & 90\% \\

\midrule

\multicolumn{3}{l}{\textbf{Joint evaluation outcomes}} \\
Tie on both dimensions & 50\% & 34\% \\
Guided preferred on both dimensions & 10\% & 24\% \\
Guided preferred for quality, tie on adherence & 16\% & 14\% \\
Guided preferred for quality, default preferred for adherence & 4\% & 0\% \\
Guided preferred for adherence, tie on quality & 0\% & 12\% \\
Guided preferred for adherence, default preferred for quality & 2\% & 0\% \\

\midrule

\multicolumn{3}{l}{\textbf{Inter-annotator agreement (3 annotators)}} \\
Krippendorff's $\alpha$ (overall quality) & 0.45 & 0.41 \\
Krippendorff's $\alpha$ (guideline adherence) & 0.25 & 0.37 \\
Pairwise $\kappa$ range (overall quality) & 0.33--0.61 & 0.38--0.48 \\
Pairwise $\kappa$ range (guideline adherence) & 0.16--0.30 & 0.37--0.40 \\

\midrule

\multicolumn{3}{l}{\textbf{Human vs. metric agreement}} \\
COMET $\kappa$ (overall quality) & 0.54 & 0.50 \\
COMETKiwi $\kappa$ (overall quality) & 0.62 & 0.54 \\
COMET $\kappa$ (guideline adherence) & 0.36 & 0.56 \\
COMETKiwi $\kappa$ (guideline adherence) & 0.50 & 0.46 \\

\bottomrule
\end{tabular}
\caption{
Summary of the human evaluation results on \rocsmt\ and \pfsmb.
Each dataset contains 50 sampled sentence pairs evaluated under the default and matching-guideline prompting conditions using outputs from \gemmafull\ and \llamafull.
Preference percentages are computed from majority-vote annotations.
Inter-annotator agreement is reported using Krippendorff's $\alpha$ and pairwise Cohen's $\kappa$.
Human vs.\ metric agreement corresponds to Cohen's $\kappa$ between majority-vote human preferences and automatic metric preferences.
Metric score differences within $\pm0.5$ points are treated as ties.}
\label{tab:human-eval-full}
\end{table*}

\end{document}